\documentclass{article}

\usepackage[preprint]{neurips_2026}

\usepackage[utf8]{inputenc} 
\usepackage[T1]{fontenc}    
\usepackage{url}            
\usepackage{booktabs}       
\usepackage{amsfonts}       
\usepackage{nicefrac}       
\usepackage{microtype}      
\usepackage[most]{tcolorbox}
\usepackage{dcolumn}
\usepackage{xcolor}         
\definecolor{darkblue}{HTML}{627EBC}
\definecolor{darkred}{HTML}{C0483A}
\usepackage[breaklinks,colorlinks=true,citecolor=darkblue,linkcolor=darkred,urlcolor=darkblue]{hyperref}

\usepackage{graphicx}
\usepackage{soul}
\usepackage{amsmath}
\usepackage{enumitem}
\usepackage{xurl}
\newcommand{\llama}{Llama-3.1-8B}
\newcommand{\gpt}{GPT-4o}
\newcommand{\supp}{Appendix} 

\tcbset{
  llmprompt/.style={
    enhanced,
    width=\linewidth, 
    colback=white,
    colframe=darkblue!60,
    coltitle=black,
    colbacktitle=darkblue!10,
    fonttitle=\bfseries\small\sffamily,
    arc=1.5mm,
    boxrule=1pt,
    top=1mm, bottom=1mm, left=1mm, right=1mm,
    boxsep=4pt,
    before skip=10pt, after skip=10pt,
    breakable,
    listing only,
    listing options={
      basicstyle=\ttfamily\small,
      breaklines=true,
      breakatwhitespace=false,
      columns=flexible,
      keepspaces=true,
      showstringspaces=false
    },
  }
}

\title{Quantifying Human-AI Synergy}

\author{%
  Christoph Riedl\\
  Khoury College of Computer Sciences\\
  Northeastern University\\
  Boston, MA 02115\\
  USA\\
  \texttt{c.riedl@northeastern.edu} \\
  \And
  Ben Weidmann\\
  Social Research Institute\\
  University College London\\
  Gower St, London, WC1E 6BT\\
  United Kingdom\\
  \texttt{b.weidmann@ucl.ac.uk}
}

\begin{document}

\maketitle

\begin{abstract}

Humans increasingly collaborate with AI, yet most research focuses on AI's net impact without modeling human-AI performance as a collaborative process. This limits our ability to understand who benefits from AI, and why. We develop a framework to explore the joint contributions of humans and AI. We separately estimate each person's individual and collaborative ability, while accounting for task difficulty. Using data from 667 people answering 396 questions alone and with one of two large language models ($11{,}008$ total answers), we find substantial synergy that cannot be explained by people simply adopting the AI's answers. Applied to a second dataset of one-way AI advice with explanations ($1{,}621$ individuals; $65{,}840$ answers), the same framework finds gains only where the AI is right and losses wherever it errs, showing that it distinguishes collaboration mechanisms rather than finding synergy by construction. We show that collaborative ability is a latent, measurable trait, distinct from individual problem-solving ability: people who perform best alone are not necessarily those who perform best with AI. Distinguishing these two dimensions reconciles whether AI equalizes or amplifies skill: AI acts simultaneously as a substitute for solo ability and a complement to collaborative ability. We find that people who are better at inferring what the AI knows collaborate more effectively with it---but are no better at solving problems when working alone---and that moment-to-moment shifts in perspective-taking predict the quality of the AI's responses. 

\end{abstract}

\section*{Introduction}

Collaboration between humans and artificial intelligence---especially large language models (LLMs)---is rapidly expanding in cognitive work \citep{bick2026rapid}. Firms are redesigning jobs around AI, educators are deciding whether and how to teach students to work with it, and policymakers are debating whether it will narrow or widen the skill gap. 
Yet what is known about human-AI collaboration is largely confined to its net effect on performance from comparing people working alone to those working with AI \citep{noy2023experimental,bansal2021does,vaccaro2024combinations,brynjolfsson2025generative,dell2026cybernetic,dell2023navigating}. 

While these productivity studies generally find gains, comparisons of mean effects leave many questions unanswered. For example, it is difficult to explain whether gains reflect genuine synergy or substitution \citep[with people simply adopting answers the AI already had;][]{steiner1972group}. Distributionally, focusing on mean performance often obscures which individuals benefit from AI and on which tasks because task difficulty, individual ability, and collaborative skill are confounded in many settings. Perhaps as a result, evidence on whether AI substitutes for or complements human skill is mixed in studies that treat skill as a single dimension \citep[e.g.,][]{brynjolfsson2025generative,otis2024uneven}. Most importantly, comparing mean performance with and without AI cannot explain \textit{why} synergy arises. An imperfect understanding of synergy may, in turn, limit the ability to design and evaluate AI models that complement and extend---rather than imitate---human capabilities \citep{srivastava2023beyond, Haupt2025AIShouldNotBeAnImitationGame, mancoridis2025potemkin, acemoglu2018artificial, riedl2024hbr}.

Here, we develop and apply a framework that moves the study of human-AI collaboration from measuring uplift to quantifying and explaining synergy. We treat human-AI interaction as a form of teamwork, building on the insight that intelligence---whether human or artificial---is inherently interactive, contextual, and collaborative \citep{minsky1985society,hutchins1995cognition,levy1997collective,riedl2021quantifying}. 
We define human-AI synergy as the performance uplift attributable to the joint human-AI process itself: gains that are not reducible to either the human's or the AI's contribution alone. This is conceptually analogous to the synergistic information atom in partial information decomposition \citep{williams2010nonnegative, riedl2026emergence}, but operationalized behaviorally (see Methods for details and empirical decomposition).
Synergy so defined includes gains from scaffolding and co-construction, but also from calibrated reliance---i.e., the joint decision of whose answer to adopt on a given task, in the sense of a transactive memory system \citep{wegner1987transactive,lewis2003measuring}. 

If synergy arises from the joint process, some people should be better at it than others. To explain why some people collaborate better with AI than others, we draw on research on human teams, where the most robust individual-level predictor of collaborative performance is Theory of Mind (ToM) \citep{riedl2021quantifying,Weidmann2021}---the capacity to represent and reason about what another agent knows and intends, which underlies social skill more broadly \citep{premack1978does,deming2017growing}. ToM allows collaborators to anticipate each other's actions, disambiguate and repair communication, and coordinate contributions during joint tasks \citep{frith2006neural,clark1996using,sebanz2006joint,tomasello2010origins}. Work in human-computer interaction likewise shows that effective collaboration with AI depends on forming accurate mental models of what the system knows and can do \citep{bansal2019updates,paleja2021utility,bansal2019beyond,gero2020mental,alipour2021improving,fugener2022cognitive,westby2023collective}. Because LLMs generate novel unpredictable output in response to user inputs, working with them plausibly benefits from similar mechanisms that support human teamwork (e.g., anticipating teammate contributions, correcting errors of understanding), making ToM a natural candidate mechanism (see \supp~\ref{sec:ToM_literature} for extended justification).

We use Bayesian Item Response Theory to model human-AI performance as a joint process. The model accounts for task difficulty, user ability, and the AI's standalone capability. We leverage partial pooling (shrinkage) to stabilize estimates and propagate uncertainty across all estimates, avoiding the pitfalls of point-estimate approximations. Crucially, our approach estimates each person's individual ability ($\theta$) and collaborative ability ($\kappa$) as separate latent traits (see Methods). This yields model-based counterfactuals that turn a single average effect into a distribution of gains across people and questions. It lets us separate synergy from substitution, and test whether synergy is a uniform boost or specific to particular user-task pairings. A bilinear extension, anchored by standalone AI benchmarks on each question, allows us to interact solo and collaborative ability with the AI's item-level capability \citep{hoff2005bilinear}. A model-free partial information decomposition provides an independent check. We apply the framework to data from 667 people who completed 396 multiple-choice questions in math, physics, and moral reasoning, first alone and then with one of two LLMs of different capability, \gpt\ and \llama~\citep[$11{,}008$ total answers;][]{Chang2025Chatbench}. To answer not only who benefits from AI but why, we score each user's prompts for signatures of ToM (see Methods). To test discriminant validity and contrast conversational collaboration with static advice-taking, we also evaluate our framework on a second dataset of non-interactive AI recommendations \citep[$1{,}621$ individuals; $65{,}840$ answers;][]{bansal2021does}.

We have three main findings. First, human-AI collaboration produces synergy. On average, individuals improve by 16-24 percentage points when working with AI, with gains persisting even with a weak AI model that scores below the typical solo human and on items the AI cannot solve alone. This suggests that people are not simply adopting AI answers. A model-free information decomposition independently confirms a positive synergistic component. By contrast, in a second dataset of one-way AI advice, human-AI teams gain nothing where the AI errs. Here, our framework exposes what mean comparisons miss: wide variation in user deference and apparent complementarity that fails to generalize beyond the specific questions used in the study.
Second, we find that the ability to collaborate with AI is a latent trait separable from individual problem-solving ability. In our setting, the collaborative ability and individual problem-solving ability both substantially contribute to human-AI joint performance. Because the two are only moderately correlated, performance rankings reshuffle when people work with AI: the best solo performers are likely, but not certain, to remain on top.
The collaboration boost is largest on the hardest problems and for the lowest-ability users, compressing the skill gap by roughly 59\%. 
Interacting both human skills with item-level AI capability reveals why: AI acts simultaneously as a substitute for solo skill and a complement to orthogonal collaboration skill.
Third, ToM predicts collaborative ability but not solo ability. Moreover, both stable individual differences and moment-to-moment variation in ToM predict the quality of AI responses, whereas individual ability does not. Read together with the lack of unstructured, idiosyncratic user-by-item dynamics, this points to social cognition as a candidate mechanism and to an asymmetry in the collaboration: the human adapts to the AI, but current models do not yet adapt to the human.
These findings turn human-AI synergy from an aggregate treatment effect into a measurable, decomposable, and explicable dimension of collective intelligence \citep{riedl2021quantifying}.

\section*{Results}
\begin{figure}[t!]
\centering
\includegraphics[width=.75\columnwidth]{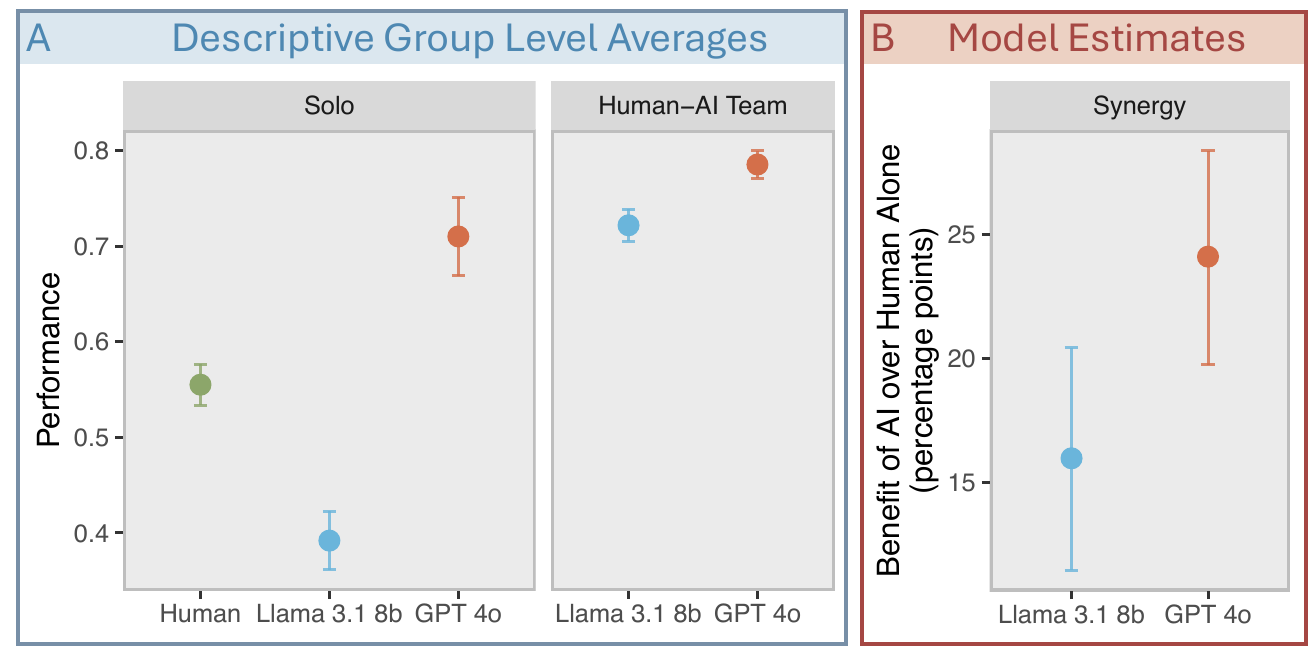}
\caption{
    \textbf{AI improves performance in human-AI teams.}
    \textbf{A} Model-free analysis of group-level averages (AI-alone accuracy). Human-AI teams perform better than human-alone. (Bars are 95\% CIs.)
    \textbf{B} The ``AI Synergy'' benchmark (percentage point change
over human alone). Average synergy with \gpt\ is nominally higher than \llama.  (IRT estimates based on Eq.~\ref{eq:model2}; bars are 95\% CrIs.)
}
\label{fig:modelFree}
\end{figure}

\paragraph{Quantifying Synergy.}\label{sec:results:benchmarking}
We begin with a model-free analysis (Figure \ref{fig:modelFree}A). Working alone, humans answer 55.5\% of questions correctly on average (based on 2,072 solo observations). \gpt\ performs better than the human average (71\% accuracy) whereas \llama\ does worse (39\%).
When working jointly with humans, even the relatively weak \llama\ model achieves substantially higher performance, significantly outperforming human-alone performance. \gpt-with-human also achieves substantial gains, outperforming \gpt-alone. However, the large difference in solo performance between \llama\ and \gpt\ shrinks dramatically. 

Next, we explore the collaborative capability of each AI model by quantifying its synergy (Figure \ref{fig:modelFree}B). We extend the results of \citet{Chang2025Chatbench} by quantifying the average impact each model has on human-AI performance \textit{while accounting for differences in task difficulty, human solo ability, and human collaborative ability}. To calculate the percentage point increase, we computed the pairwise difference in probabilities between the AI-assisted and baseline conditions directly across posterior draws. A key advantage of working directly with the posterior draws is that it propagates the joint uncertainty of all parameter estimates, avoiding the pitfalls of point-estimate approximations. We find that \llama\ boosts performance by 16 percentage points on average, while \gpt\ boosts performance by 24 percentage points. Credible intervals overlap, suggesting no clear advantage for either model.\footnote{As a robustness test (\supp~\ref{sec:learning}), we estimate a three-phase model that uses an additional 3,056 observations from a second ``solo first'' condition and separately identifies learning or fatigue effects. It confirms the main results are not driven by the fact that the with-AI condition always followed the solo condition and, if anything, slightly underestimate AI’s true benefit.}
To explore the mechanism behind this complementarity, we quantify shifts from answering incorrectly alone to answering correctly with the help of AI via posterior predictive counterfactuals. On average, users switch from wrong answers to right answers on 25\% of the tasks when working with \llama~and 29\% when working with \gpt. Harmful switches from right to wrong are rare (12.3\% and 8.6\%, respectively) and, per opportunity to be harmed, independent of ability. At the individual level, item-level harm is offset by gains: no user (0 of 667) shows a credibly negative expected boost with either AI model (see \supp~\ref{sec:complementarity} for method details and additional results, including a breakdown by task difficulty and domain). 

\paragraph{Synergy Beyond Direct Substitution.} 
A separate question is whether the benefits of working with AI reflect genuine complementarity---gains from scaffolding, clarification, and co-construction---or substitution, where the AI provides the correct answer and the user adopts it. We provide several tests. First, controlling for an exogenous measure of each question's AI standalone accuracy, user-level collaborative ability retains 73\% of its standard deviation (53\% of its variance). We also find that users improve over their solo baseline even on questions the AI could not solve alone ($\beta = 0.80$ [95\% CrI: 0.59, 1.01]). In information-theoretic terms, this bounds the unique contribution that AI makes (\citealp{riedl2026emergence}; i.e., $I(A;Y) \approx 0$ and thus $\mathrm{Un}(A) \approx 0$). Hence, gains are unlikely to be substitution.
Second, we split problems into ``AI-solvable'' and ``AI-unsolvable'' based on the AI solo performance. We find that collaborative ability (see next section for details) estimated separately on both sets is strongly positively correlated ($\rho = 0.80$ [95\% CrI: 0.53, 0.96]). This points to a unified collaboration skill---active elicitation and error-checking, and appropriately calibrated reliance on AI output \citep{Caplin2025ABC}---not passive adoption (\supp~\ref{sec:substitution}). 
Third, a model-free partial information decomposition on the answer-first arm \citep{riedl2026emergence}, where participants' own and final answers are observed on the same questions, recovers a credibly positive synergistic atom, with its share of total information rising on harder questions (Figure \ref{fig:PID}). The decomposition is inconsistent with pure substitution, under which final answers would track the AI's and information would load onto the AI-unique rather than the human-unique atom (\supp~\ref{sec:PID}).

\paragraph{Discriminant Validity: Interactive Synergy vs.~Static Advice-Taking.} 
Does our framework recover synergy on AI-unsolvable items by construction, or does it genuinely distinguish dialogic collaboration from passive advice-taking? To test this, we apply our item-level IRT approach to a second dataset from a non-interactive setting \citep{bansal2021does}. Here, users receive static AI advice with explanations but without dialogue ($N=1{,}621$ across two sentiment and one logical-reasoning task; see \supp~\ref{si:bansal}). (Because users were randomized between solo and joint AI conditions, this design serves as a test of the item-level substitution mechanism rather than within-user $\theta$--$\kappa$ separation or ToM that we analyze in the sections below.) Although human-AI teams outperform both solo humans and the AI on average as originally reported, our framework reveals a starkly different mechanism from conversational LLMs: gains in the static setting come almost entirely from users deferring to the AI when it is right, but they realize only 21--38\% of the gain available under oracle selection. Crucially, unlike in conversational interaction, joint performance falls below solo human performance on every item the AI gets wrong. This provides a clear negative control, confirming that the gains we observe on AI-unsolvable items in dialogic interaction are a substantive property of back-and-forth collaboration rather than an artifact of the IRT estimator. Moreover, a crossed random-effects model fitted to data from the joint condition alone (Eq.~\ref{eq:bansal}), with item effects, random intercepts for each individual (overall judgment), and a random slope on whether the AI's advice was correct (deference), reveals that participants vary up to nearly as much in their deference to the AI as in their judgment (SD 0.45–0.73 vs.\ 0.73–0.90, logit scale). Individual accuracy with AI thus reflects reliance as well as skill, and would be expected to shift with the AI's error rate, a distinction that comparisons of average gains cannot make. It is also counter to the synergy effect we find in our main dataset.  That is, complementarity in a selection setting depends on how accurate the AI happens to be, not on how well the team collaborates. We test this specifically by re-estimating the headline complementarity results using our modeling framework but treating items as random samples. Using this estimation approach, the finding that human-AI teams beat their best member generalizes on the two classification tasks but \textit{not} on the logical reasoning task (LSAT).

\paragraph{Within-User Ability Decomposition.}
To test whether working alone and with AI are separately estimable latent abilities we fit the model from Eq.~\ref{eq:model1} and a nested, simpler model that estimates a single latent ability for each user. We perform Bayesian model comparison with leave-one-out cross-validation to evaluate how well each model predicts unseen data by estimating the expected log predictive density (ELPD) for held-out observations. We find the full model fits the data substantially better ($\Delta$ELPD = 50.9, SE = 10.2). Overall model convergence is excellent (all $\hat{R}$ are 1.00 showing no evidence of chains mixing poorly) and both Bulk ESS and Tail ESS are large (all $> 1{,}500$, some $> 4{,}000$) indicating very good sampling efficiency and stable posterior estimates. 

The model comparison decisively favors the two-ability specification, which is consistent with condition-specific latent abilities---one reflecting users' ability to work alone ($\theta$), and one capturing their ability to work with AI ($\kappa$)---supporting the notion that $\theta \neq \kappa$. Because $\theta$ and $\kappa$ are allowed to covary freely, any collaboration-relevant skill that is linearly predictable from solo ability is attributed to the shared component; the share of latent variance in $\kappa$ orthogonal to $\theta$ is therefore a lower bound on collaboration-specific variance. Even using the highest correlation we estimate---in the subsample with six solo items, where $\theta$ is measured most precisely ($\rho = 0.64$ [0.48, 0.79])---at least 38\% of the latent variance in $\kappa$ (the lower 95\% bound) is not predictable from solo ability.
Note that this is a claim about the population distribution of abilities---estimated from the entire dataset of 667 individuals---and not about any individual's specific ability rank. Several robustness checks support these conclusions: the $\theta \neq \kappa$ separation replicates on the subset of users with six solo observations giving more precise ability estimates; solo-ability estimates are reliable and stable across the first and second solo phase; and our results are insensitive to prior specification (\supp~\ref{sec:robustness} and~\ref{sec:priors}).


\paragraph{Non-Additive Synergy.} We perform several tests to answer whether human-AI synergy is additive or non-linear. First, we test for non-additive synergy directly by asking whether the AI boost bends with item difficulty in a user-specific way. That is, we introduce a $\text{user} \times \text{difficulty}$ interaction term (three bins). We target difficulty rather than domain or the raw $\text{user} \times \text{item}$ residual because difficulty varies in the dataset across a user's items (domain is nested within user, and the cell-level residual is confounded with Bernoulli noise). We find no detectable user-specific difficulty-dependent non-additive synergy ($\Delta$ELPD = -1.7, SE = 1.2; adding the interaction does not reveal hidden non-linearity $\Delta R^2_{MZ} = 0.003$). A simulation-based power analysis using our full estimation pipeline shows that we would have detected user-by-difficulty non-linear synergy exceeding 16\% of the additive-synergy variance (interaction SD $0.63$ logits) with $\geq$80\% power (\supp~\ref{sec:power}). Hence, the absence of such structure bounds any non-linearity to a small fraction of the additive effect rather than leaving it unquantified. 

Second, we estimated a 2PL \citep{baker2004item} extension of the model that allows item-specific discrimination parameters, permitting the mapping from latent ability to response probability to vary across items---the simplest form of $\text{user} \times \text{item}$ interaction identifiable with single-trial Bernoulli data. The standard deviation of log-discrimination was small (0.23 [95\% CrI: 0.02, 0.46]) and LOO-IC comparison yielded no reliable improvement over the additive model ($\Delta$ELPD = 13.7, SE = 13.6), again indicating no reliable non-additive structure.

Third, we decompose performance into a structural signal and stochastic noise (Eq.~\ref{eq:synergy_decomp}). We recover 45.7\% of the total variance as explainable signal ($R^2_{MZ} = 0.457$); crucially, we find that the remaining variance is consistent with aleatoric noise rather than recoverable structure. The residual dispersion statistic ($\phi = 0.658$) is consistent with the absence of systematic non-linearities. Furthermore, variance analysis shows that the model fits collaborative data ($\sigma^2_{\text{joint}} = 0.63$) nearly as robustly as the solo baseline ($\sigma^2_{\text{solo}} = 0.51$). Within this robust structure, we find that the additive boost parameter is a primary driver of performance, accounting for 42.5\% of the reliable signal variance. 

These diagnostics support a central conclusion: 
Human-AI synergy is free of idiosyncratic user $\times$ item dynamics; the non-additivity that exists is low-dimensional, running through collaboration skill and AI capability.
This establishes AI collaboration ability as a first-order latent trait---distinct from but comparable in magnitude to individual ability. It also supports the proposed framework as a high-fidelity instrument for benchmarking human-AI synergy. 
Predominant additivity at the latent level is not in tension with the presence of synergistic information in the information-theoretic sense: any aggregation rule under which the pair succeeds when either partner would have (an OR-like channel) carries substantial synergistic information under partial information decomposition, and a model-free decomposition on the answer-first subset confirms a credibly positive synergistic atom (\supp~\ref{sec:PID}). The dispersion and interaction tests establish that the mechanism generating this information is a stable, generalized channel ($\kappa$) rather than idiosyncratic user-by-item dynamics. That is, synergistic information is produced by a predominantly additive, low-dimensional process \citep{riedl2026emergence}.
One interpretation consistent with these bounds is that only one agent---the human---actively adapts to the other while fully bilateral mental modeling would be expected to introduce interaction-level variance that exceeds, rather than falls below, the additive baseline.

\begin{figure*}[t!]
\centering
\includegraphics[width=1\textwidth]{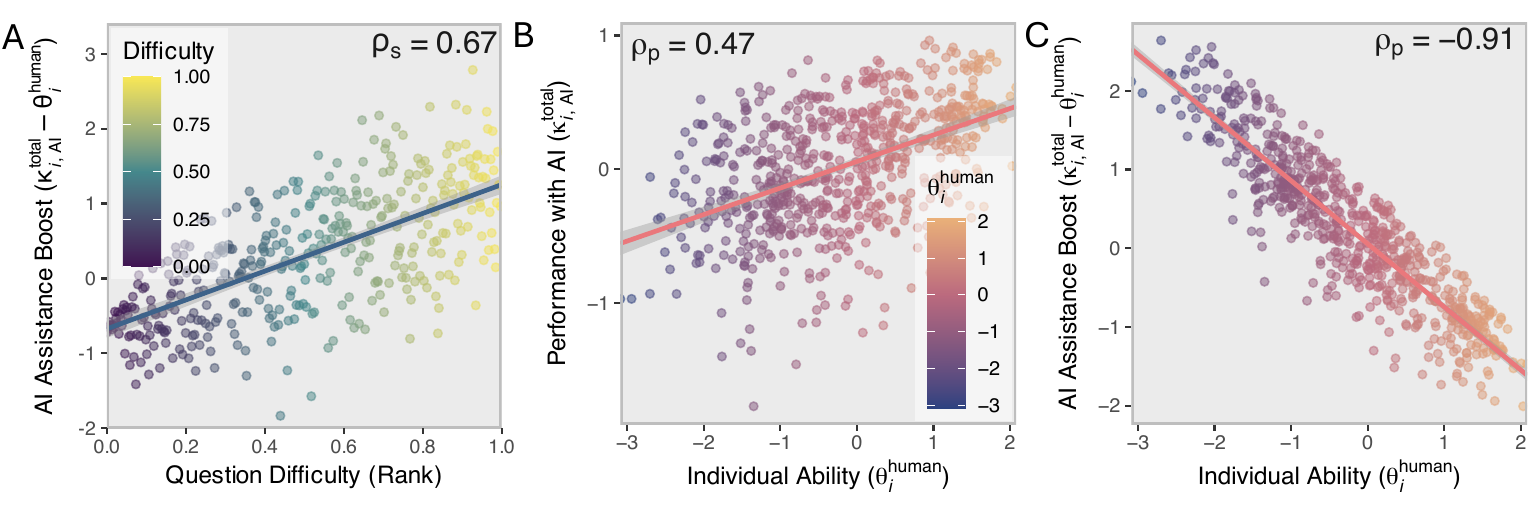}
\caption{
    \textbf{Who benefits from AI and on which task?}
    \textbf{A} AI boost is largest on the hardest questions.
    \textbf{B} In the ``with AI'' condition, higher-ability users perform best in absolute terms.
    \textbf{C} AI boost is highest for lower-ability users. Note that the AI boost can be limited by ceiling effects for high-ability users. 
    Points are posterior means with partial pooling; 
    lines are population-level best-fit linear regressions with 95\% confidence bands. 
}
\label{fig:IRT}
\end{figure*}

\paragraph{Who benefits from AI?}
Having quantified \textit{average} synergy we now decompose these effects. 
First, we estimate the impact of AI on questions of differing difficulty.
We find that  more difficult tasks (by solo accuracy) benefit more from AI collaboration (Figure \ref{fig:IRT}A). This is consistent with theories of human-machine complementarity suggesting that AI provides greater additive benefit where human performance is weakest \citep{autor2015there,acemoglu2018artificial}. 
In information-decomposition terms, this is also the regime in which redundant information is structurally starved: as solo accuracy approaches the floor, the human's answer carries little information about the outcome, and with it the redundancy bound shrinks as well \citep{riedl2026emergence}. 
Uplift on the hardest questions therefore cannot reflect information both parties already held; it must be unique to the AI or synergistic (\supp~\ref{sec:substitution} shows it persists on items the AI cannot solve alone; see also \supp~\ref{sec:PID}).

\begin{figure*}[t!]
\centering
\includegraphics[width=1\textwidth]{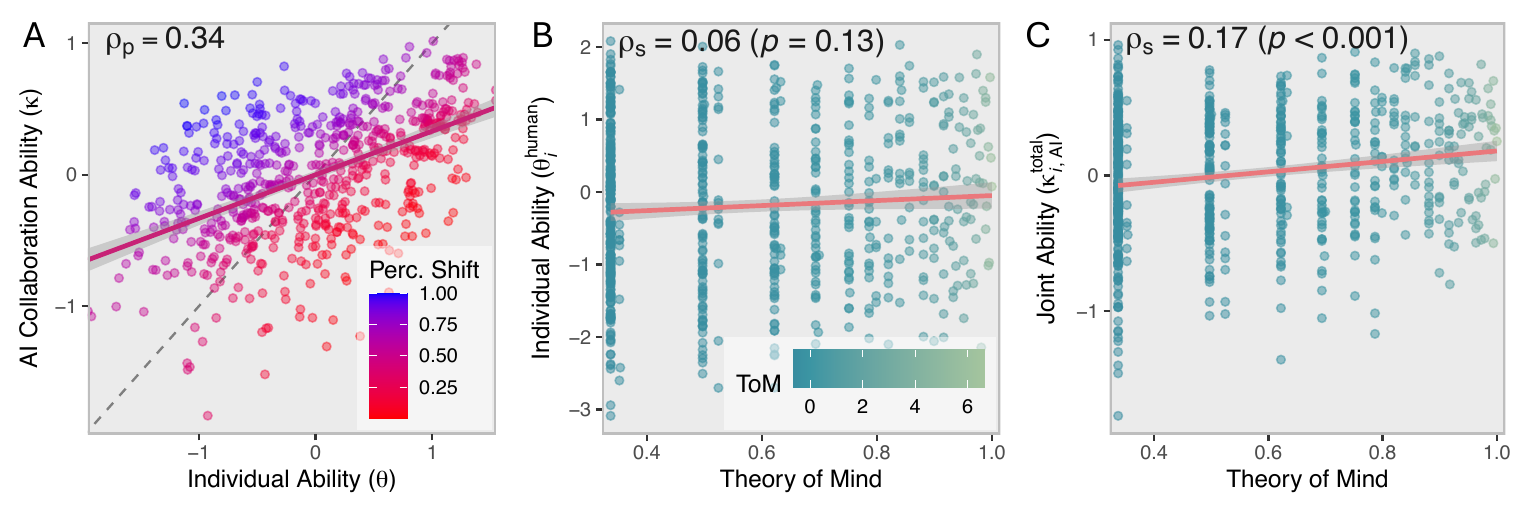}
\caption{
    \textbf{Disentangling individual and collaborative ability: Theory of Mind predicts performance during collaboration with AI but not when working alone.}
    \textbf{A} Individual and collaborative ability are moderately correlated ($\rho_p=0.34$ [95\% CrI: 0.15, 0.54]). Color shows normalized shift in rank (estimates based on Eq.~\ref{eq:model1} with correlated covariance).
    \textbf{B} ToM is not significantly correlated with individual solo ability.
    \textbf{C} ToM is significantly correlated with the AI collaboration ability.
    Abilities are IRT estimates from Eq.~\ref{eq:model1}. ToM is raw data aggregated to the user level. 
    Points are posterior means with partial pooling; 
    lines are population-level best-fit linear regressions with 95\% confidence bands. 
}
\label{fig:ToM}
\end{figure*}

Next, we test whether lower- or higher-ability users benefit more from AI (Figure \ref{fig:IRT}B). We find a strong positive correlation between individual solo performance ($\theta_{i}^{human}$) and joint performance with AI ($\kappa_{i, \text{AI}}^{\text{total}}$), shown in Figure \ref{fig:IRT}B. However, the AI boost estimates suggest that lower-ability users \textit{gain more} than their higher-ability peers (Figure \ref{fig:IRT}C). Notice that these results are population-level patterns with correlations and variance components identified from the full sample. These results are thus not dependent on the precision with which any single user's rank is estimated (the individual points in Figures \ref{fig:IRT} \& \ref{fig:ToM} are posterior means illustrating population patterns rather than claims about a specific user's rank).

This result mirrors prior findings that for closed-form tasks, AI reduces---but does not fully eliminate---output inequality across initial skill levels \citep[e.g.,][]{noy2023experimental,dell2023navigating}.\footnote{This result is partly driven by the fact that high-ability users have less room for improvement (i.e., a ceiling effect).} 
We also test whether \gpt\ or \llama\ is better at complementing users at different solo ability levels but find this is not the case (see \supp~\ref{sec:robustness}).

\paragraph{Decomposing Synergy: Equalization and Distinctness.} 
Our Bayesian formulation allows us to further disentangle the relationship between individual and AI collaboration ability by relaxing the independence assumption of the model (Figure \ref{fig:ToM}A). This sheds more light on the two complementary phenomena mentioned above. First, we find an equalizing effect through variance compression. Comparing the latent trait distributions, we find that AI collaboration significantly reduces performance variance ($\sigma_{solo}^2 = 1.39$ vs.~$\sigma_{joint}^2 = 0.58$). 
This explains the strong negative correlation between solo ability and AI boost (lifting up lower-ability users while simultaneously imposing a ceiling on high-ability users) and confirms that AI compresses the skill gap. AI benefits low-ability users disproportionately and the spread shrinks by 59\%. 

Second, we find a positive, moderate correlation between individual and collaboration ability ($\rho_p = 0.34$ [95\% CI: 0.15, 0.54]). Pure automation would imply $\rho \approx 0$; pure amplification of solo skill would imply $\rho \approx 1$ with variance expansion. Concretely, this means performance rankings shuffle when users move from solo to collaborative settings (Figure \ref{fig:ToM}A). The top solo performers are \textit{likely} but not \textit{certain} to remain on top, because a largely independent ability now co-determines outcomes. While higher-ability solo performers retain an absolute advantage, AI does not act as a multiplier for domain experts: individual ability leaves most of the variance in joint performance (88.4\% here; at least 38\% in our six-item subsample) to distinct collaborative skill.

\paragraph{Structural Mechanism of Synergy.} 
To probe the mechanism behind human-AI synergy, we examine how distinct human skills combine with item-level AI and task characteristics. While unstructured cell-level user $\times$ item interactions are not identifiable with single-trial Bernoulli data (see above), structured interactions are: if the interaction runs through a small number of latent user and item attributes, each measured in a separate experimental arm, it is identified from the pattern across many user-item cells. Building on the bilinear mixed-effects framework of \citet{hoff2005bilinear}, we develop a three-arm specification that anchors latent item attributes using standalone human and AI trials to estimate $2 \times 2$ multiplicative complementarity between two individual skills ($\theta$ and $\kappa$) and two task attributes (model-item-specific AI accuracy $\eta_{jm}$ and human difficulty $d_j$; see \supp~\ref{si:mechanism}).
Complementarity requires humans and AI to cover each other's weaknesses \citep{bansal2021does,steyvers2022bayesian}; unidimensional models cannot represent this, but our $2 \times 2$ model does so explicitly.
Unlike Eq.~\ref{eq:model1}, where $\theta$ and $\kappa$ are estimated as correlated traits ($\rho = 0.34$), here we decompose collaboration ability into its projection on solo skill ($\theta_i$) and solo-orthogonal collaboration skill ($\kappa^\perp_i \perp \theta_i$). This cleanly separates domain expertise from collaboration skill and avoids collinearity in the interaction terms.
Adding the $2 \times 2$ latent user-item interactions modestly improves out-of-sample predictive fit on the collaboration-trial data over the nested additive baseline  (PSIS-LOO $\Delta$ELPD = 5.8, SE = 4.8). A joint posterior Wald test confirms that the interaction block is highly statistically significant ($\chi^2(4) = 13.34, p = 0.0097$).

Working with AI roughly halves the payoff to solo ability: a unit of $\theta_i$ is worth $b_1 = 0.46$ log-odds in the joint task versus $1.00$ alone (95\% CrI $[0.24, 0.77]$; $\Pr(b_1 < 1) = 0.997$). Nor does solo ability pay off more when the AI is strong on an item ($b_3 = 0.00$ $[-0.04, 0.04]$). Collaboration skill behaves differently. It has a large main effect ($b_0 = 0.65$ $[0.43, 0.83]$), and it pays off more on items the AI tends to get right: per logit of item-level AI accuracy, a one-SD increase in $\kappa^\perp_i$ adds $b_5 = 0.07$ $[0.02, 0.11]$ log-odds of a correct joint answer ($\Pr(b_5 > 0) = 0.997$; one coefficient for both models, net of human-difficulty interactions $b_7$ = -0.15 [-0.40, 0.06]; $b_8$ = 0.21 [-0.07, 0.51]). The per-logit effect is small, but AI accuracy varies widely across items (SD $7.0$ logits for \gpt, $3.8$ for \llama), so a one-SD easier item for the AI is worth roughly $0.5$ log-odds (\gpt) or $0.3$ (\llama) to a +1\,SD collaborator. Collaboration skill amplifies AI capability more than solo skill does ($b_5 - b_3 = +0.065$ $[0.000, 0.127]$, $\Pr(b_5 > b_3) = 0.976$; this direct comparison is valid because $b_3$ and $b_5$ multiply the same $\eta_{jm}$).
This pattern is inconsistent with collaboration skill consisting mainly of rescuing standalone AI failures (which would imply $b_5 \le 0$).
The bilinear model also allows us to speak to synergy directly. On user-item pairs that both partners are likely to get wrong alone (posterior-mean $\Pr(\text{correct}) < 0.5$ for both), human-AI teams beat item-level deference to the more accurate partner ($+4.6\text{ pp}$ $[0.9, 8.1]$, $\Pr(>0) = 0.995$), while high-$\kappa^\perp$ collaborators ($+1\text{ SD}$) further exceed the truth-wins OR ceiling ($+7.0\text{ pp}$ $[0.2, 13.5]$, $\Pr(>0) = 0.978$ pooled across models; $+9.0\text{ pp}$ $[2.4, 15.7]$, $\Pr(>0) = 0.995$ with Llama). This is a conservative lower bound on interactive gains beyond pure answer selection, such as co-construction or scaffolding.
AI thus flattens the returns to solo skill but amplifies the returns to orthogonal collaboration skill, consistent with calibrated reliance and with gains that arise only through interaction.



\paragraph{Theory of Mind.} To explain the observed human-AI synergy, we draw on established theories from human-human collaboration, particularly Theory of Mind. ToM refers to the capacity to represent and reason about others' mental states \citep{premack1978does} and plays a crucial role in human-human interaction \citep[][see \supp~\ref{sec:ToM_literature} for additional background on ToM]{nickerson1999hidden,lewis2003measuring}. As measured here, ToM is the tendency to model the AI's capability and informational state, and tailor prompts to it. It is distinct from prompt effort (turns, characters, edit ratio) and from surface prompt quality (concreteness, readability, spelling), both of which we control for directly.
We collect task-level ToM measures from the users' portion of the dialogue, using an LLM as a research assistant \citep[LMRA,][see \supp~\ref{sec:ToM_LMRA}]{eloundou2024first} and follow best practices for using LLMs in this way \citep{li2025generation}. We performed extensive robustness tests to validate construct validity including (a) collecting data from a variety of leading models, (b) exploring robustness to the specific instruction set in the prompt, (c) testing for order bias in the prompt, and (d) comparing absolute vs.~relative judgments. We find excellent inter-coder reliability among the different LLM ``raters''. We also collected gold standard human ratings for a stratified random sample of 120 dialogues to validate the approach of studying dialogue-level ToM signatures in human-AI collaboration. The ensemble LMRA agreement is excellent (ICC(2,k) = 0.90; a single human rater is only 0.13). One LMRA corresponds to about five human raters. We find substantial agreement between the gold standard and LMRA ratings (ICC(2) = 0.42; ICC(2,k) = 0.59 [95\% CI: 0.42, 0.72]), outperforming a single human rater and approaching the consistency of the aggregated human consensus (see \supp~\ref{sec:ToM_LMRA}, \ref{sec:Human_gold}, \ref{sec:LMRA_ToM_construct}).

We find that ToM differentially predicts performance in the AI condition (ToM $\times$ Joint interaction, Model from Eq.~\ref{eq:model3}: 0.65 [95\% CrI: 0.02, 1.26]); in the solo condition, ToM is not credibly associated with performance (0.40 [95\% CrI: -0.16, 0.99]) (ToM is even more strongly correlated with the solo-orthogonal collaboration ability ($\kappa^\perp$) from the bilinear model: $\rho_s = 0.22, p = 2.4 \times 10^{-8}$).\footnote{\label{fn:ToM_robustness}We find substantively similar results using a different LLM to assess ToM signatures in users' prompts. Specifically, using Google's Gemini 2.5 Pro we find ToM is \textit{not} credibly associated with accuracy in the solo condition  (0.34 [95\% CrI: -0.16, 0.84]) but is credibly strong in the joint condition (ToM $\times$ Joint: 0.58 [95\% CrI: 0.03, 1.12]).} 
When we re-estimate model Eq.~\ref{eq:model3} using a narrower operationalization of ToM (\supp~\ref{sec:ToM_LMRA}), the ToM $\times$ joint condition interaction gains strength and significance ($\beta = 1.53$ [95\% CrI: 0.48, 2.57]). The differential pattern---ToM predicts collaborative performance but not solo performance---replicates exactly ($\beta = 0.50$ [95\% CrI: -0.49, 1.51]). Notably, we find that the average fixed-effect benefit of AI collaboration is fully moderated by Theory of Mind: at zero ToM, the AI condition provides no measurable benefit ($\beta = -0.07$ [95\% CrI: -0.84, 0.70]). This is consistent with ToM being more than a correlate of performance, and with ToM acting as a channel through which AI collaboration benefits emerge.
Taken together, these findings are consistent with the idea that Theory of Mind---belief tracking, perspective taking, goal inference etc.---captures a user-specific form of social-cognitive skill that may help explain human-AI synergy.


\begin{table}[!t]
\centering
\begin{scriptsize}
\begin{tabular}{@{\extracolsep{2pt}}l D{.}{.}{3.3} D{.}{.}{3.3} D{.}{.}{3.3} D{.}{.}{4.5}}
\toprule
Dependent Variable: 	& \multicolumn{4}{c}{Quality of AI Responses} \\
\midrule
		& \multicolumn{3}{c}{OLS} & \multicolumn{1}{c}{Bayesian} \\
 \cline{2-4} \cline{5-5} \\[-5pt]
            & \multicolumn{1}{c}{(1)} & \multicolumn{1}{c}{(2)} & \multicolumn{1}{c}{(3)} & \multicolumn{1}{c}{(4)} \\
\midrule
User ToM (LOO mean)               & 0.16^{**} & 0.24^{***}      & 0.27^{***}      & 0.31^{*}       \\
                                  & (0.05)    & (0.05)          & (0.05)          & \multicolumn{1}{c}{[ 0.22;  0.39]} \\
Solo Ability                      &           & 0.01            & 0.01            & -0.01          \\
                                  &           & (0.03)          & (0.03)          & \multicolumn{1}{c}{[-0.06;  0.05]} \\
Within-User ToM Dev.              &           &                 & 0.09^{*}        & 0.10^{*}       \\
                                  &           &                 & (0.04)          & \multicolumn{1}{c}{[ 0.05;  0.15]} \\
\underline{Controls}\\
\quad Effort (Turns)              &           & -0.17^{***}     & -0.19^{***}     & -0.17^{*}      \\
                                  &           & (0.03)          & (0.04)          & \multicolumn{1}{c}{[-0.21; -0.12]} \\
\quad Effort (Total Chars)        &           & 0.30^{***}      & 0.29^{***}      & 0.21^{*}       \\
                                  &           & (0.06)          & (0.06)          & \multicolumn{1}{c}{[ 0.15;  0.26]} \\
\quad Effort (Edit Ratio)         &           & -0.71^{***}     & -0.72^{***}     & -0.76^{*}      \\
                                  &           & (0.04)          & (0.04)          & \multicolumn{1}{c}{[-0.81; -0.72]} \\
\quad Quality (Concreteness)      &           & -0.00           & -0.01           & -0.05^{*}      \\
                                  &           & (0.04)          & (0.04)          & \multicolumn{1}{c}{[-0.09; -0.01]} \\
\quad Quality (SMOG Complexity)   &           & 0.25^{***}      & 0.25^{***}      & 0.10^{*}       \\
                                  &           & (0.04)          & (0.04)          & \multicolumn{1}{c}{[ 0.06;  0.15]} \\
\quad Quality (Spelling Mistakes) &           & -0.06^{\dagger} & -0.06^{\dagger} & -0.06^{*}      \\
                                  &           & (0.03)          & (0.03)          & \multicolumn{1}{c}{[-0.10; -0.02]} \\
\midrule
N                     & \multicolumn{1}{c}{5,868} & \multicolumn{1}{c}{5,868} & \multicolumn{1}{c}{5,868} & \multicolumn{1}{c}{5,868} \\
Question Fixed Eff.             & \multicolumn{1}{c}{396}   & \multicolumn{1}{c}{396}   & \multicolumn{1}{c}{396}  & \multicolumn{1}{c}{396}   \\
Adj.\ R$^2$           & \multicolumn{1}{c}{0.23}  & \multicolumn{1}{c}{0.41}  & \multicolumn{1}{c}{0.41}  \\

\bottomrule
\multicolumn{4}{@{}l@{}}{\tiny{$^{***}p<0.001$; $^{**}p<0.01$; $^{*}p<0.05$; $^\dagger p<0.1$}}
\end{tabular}
\end{scriptsize}
\caption{\textbf{Theory of Mind, both as user-level trait and within-user variation, predicts higher quality AI responses.} 
OLS models show standard errors clustered by individual.
Bayesian model shows 95\% credible interval.
}
\label{table:ToM_regression_2}
\end{table}


The above analysis raises an interesting question: is ToM a fixed user-level trait \textit{associated} with an individual's latent ability to collaborate---or can we view ToM as the \textit{direct mechanism} through which collaboration synergy unfolds \citep{riedl2026gbs}? These are testable questions: a direct relationship implies that a user's ToM would influence the quality of human-AI dialogues and, more specifically, the quality of the AI responses. Moreover, a direct relationship would suggest the possibility that dynamic within-user changes in ToM expression may influence the quality of AI responses. 

To explore these questions we examine the relationship between ToM and AI response quality (Table \ref{table:ToM_regression_2}; for LMRA-based quality scores see \supp~\ref{sec:AI_Quality}). Using variance decomposition (\supp~\ref{supp:ToM_method}) we find that trait-level ToM predicts AI response quality, while solo ability does not (Models 1-2), mirroring findings from human teamwork that synergy depends on ToM rather than individual problem-solving ability \citep{Weidmann2021}. Notably, the association strengthens rather than attenuates when we control for prompting effort and proxies for prompt quality, which is the opposite of what a pure prompt-specification account would predict; solo ability remains null throughout. Furthermore, within-user, question-by-question variation in ToM predicts response quality (Model 3), with consistent effects in the hierarchical Bayesian specification (Model 4). 

A natural alternative is that ToM signatures reflect prior experience with LLMs rather than a general social-cognitive capacity. Two features of our results contradict this account. First, the within-user, question-to-question variation in ToM predicts AI response quality net of each user’s own mean. A stable trait such as AI literacy cannot produce this pattern. Second, because AI responses depend only on the user’s prompt, experience can affect response quality only through the prompt's content. Yet the effect of ToM strengthens when we add controls for prompt effort and surface quality.


It is important to note a structural limit on what these tests can establish: ToM is scored from users' own prompts, so a well-specified prompt might yield a better AI response whether or not any mentalizing occurred. That said, our controls absorb the most obvious prompt-related confounds (i.e., we control for proxies of ``effort'' and ``quality''). We also note that our user-level ToM measure is a leave-one-out mean (\supp~\ref{supp:ToM_method}): a user's trait score for question $j$ is computed from their prompts on all \textit{other} questions, so the trait measure and the scored response share no text. Notwithstanding these mitigations, we emphasize that we cannot cleanly identify the causal effect of ToM and that stable user characteristics correlated with both prompting skill and mentalizing remain a plausible alternative account for our results. 
\section*{Discussion}
Our research defines human-AI synergy as the performance uplift attributable to the joint human-AI process---gains from elicitation, error-checking, and calibrated reliance that neither party would achieve alone and that are not reducible to the AI supplying an answer the user adopts \citep{riedl2026emergence}. We find such synergy: AI collaboration raises performance by 16-24 percentage points over solo baselines and users improve even on items the AI cannot solve alone. 

The uplift decomposes into a stable collaborative ability ($\kappa$) that is only moderately correlated with individual ability ($\theta$) rather than into unstructured user-by-item dynamics. Overall, the boost is largest on the hardest items and for the lowest-ability users: it compresses the performance gap between the best and worst users by 59\% without eliminating it, and no user shows a credibly negative expected boost under either model. Crucially, treating skill as two-dimensional reconciles the debate over whether AI is a skill equalizer or amplifier: it is both simultaneously. AI rotates inequality across orthogonal dimensions. It compresses returns to solo ability ($\theta$) via substitution while amplifying returns to collaboration skill ($\kappa^{\perp}$) on tasks where AI capability ($\eta$) is higher. This matches Deming's \citeyearpar{deming2017growing} task-trade logic---where social skill lowers coordination friction, making partner capability complement collaborative rather than solo skill---and extends \citet{Caplin2025ABC} from calibrated AI to conversational partners whose reliability users must infer. Stronger AI models will raise the payoff to collaboration skill especially in settings that span diverse tasks.

The claim of human-AI synergy does not hinge on any single test: the three alternative sources of uplift---substitution, shared knowledge, and human-alone ability \citep{riedl2026emergence}---are each excluded by multiple lines of evidence (\supp~\ref{sec:robustness}, \ref{sec:complementarity}, and \ref{sec:substitution}). A model-free information decomposition on the answer-first arm, which shares no assumptions with the IRT model, independently confirms a positive synergistic component (\supp~\ref{sec:PID}). Furthermore, applying our framework to a second dataset of one-way AI advice with explanations \citep{bansal2021does} provides a negative control. Human-AI teams beat both solo baselines on the questions studied, but their gains come almost entirely from deferring to the AI when it is right. Here, we shed light on what simple comparisons miss: Individuals differ substantially in how much they defer to the AI, and when items are modeled as representing an underlying task, prior results do not generalize cleanly. Our framework thus does not produce synergy by construction, and by modeling item and individual effects rather than aggregate accuracy, it reveals when results generalize. Because three solo items yield limited precision for any single user's rank, we make no claims about individual users; our substantive conclusions---the separability of $\theta$ and $\kappa$, trait-based nature of synergy, variance compression, and the ToM associations---are distribution-level quantities (variance components, correlations, and population slopes) that are identified from the full sample, are insensitive to the reliability of any individual's estimate. They are stable across prior specification (\supp~\ref{sec:priors}), in a subsample with twice as many solo items (\supp~\ref{sec:robustness}), and in a three-phase model that separates AI assistance from order effects (\supp~\ref{sec:learning}).


Who collaborates well is predicted by Theory of Mind. Users who better model the AI's informational state perform better with AI but gain no advantage alone. Both their trait-level and moment-to-moment ToM predict the quality of AI responses, whereas solo ability does not. The association strengthens rather than attenuates when we include controls for prompting effort and prompt quality. This extends a central result in collective intelligence—that ToM predicts team but not individual performance \citep{riedl2021quantifying,Weidmann2021}---to teams in which one member is a language model. Response quality is not an inherent property of the model alone but emerges from the interaction between human reasoning and AI capabilities. The pattern is consistent with ToM being more than an incidental correlate of performance. ToM may play a role in this regard, potentially acting as a cognitive-behavioral interface that supports the emergence of this synergy \citep[cf.][]{riedl2026gbs}. But we note that our design is observational and does not definitively establish ToM as a causal channel through which AI benefits emerge. 

That said, we build on evidence that ToM is a dynamic construct that can vary within users \citep{qureshi2010executive,Schneider2012Cognitive,Lin2010Reflexively} and find that within-user differences in ToM are associated with the quality of human-AI dialogue. 
Future work could explore whether variation in a user's ToM triggers or suppresses ToM-relevant circuits inside LLMs \citep{prakash2025language,riedl2026emergence}, which could inspire joint probes of both users and models. The insight that synergy is predominantly additive even though ToM varies dynamically supports the idea that current LLMs fail to fully support users as they modulate their behavior---for example based on the affordances of the task. This suggests training AI models for ToM-like responsiveness to users' cognitive states, rather than higher standalone accuracy, as a design target \citep{buckner2024deep}. 
A counterfactual decomposition of harmful switches points the same way: with the more accurate model, the majority of residual harm occurs despite correct advice. As as AI advice quality improves, what remains to be fixed is the interaction itself (\supp~\ref{sec:complementarity}).
Furthermore, multiplicative returns are currently confined to orthogonal collaboration skill. This balance could shift toward broader inequality-amplifying dynamics as AI models become more responsive to users' cognitive states. It may also already differ in settings that our benchmark does not capture, such as open-ended tasks without performance ceilings or longer iterative collaborations in which effects compound across turns.

Our research provides a unified approach to the key analytical challenges in quantifying human-AI synergy: it separates individual from collaborative ability, controls for task difficulty, distinguishes additive from non-linear synergy, benchmarks different AI models, and---by incorporating exogenous measures of AI standalone capability---enables researchers to bound the contribution of direct answer substitution versus emergent collaborative gains. By providing a methodological framework this paper lays a foundation for understanding and optimizing human-AI collaboration that moves beyond simple performance comparisons toward theoretical explanations of when, why, and how humans and AI systems can achieve synergy. 

The comparison across two AI models illustrates why this axis is distinct from standalone leaderboards: a 32-point gap in AI solo accuracy between \llama\ and \gpt\ shrinks to roughly six points once a human is in the loop. Hence, a model that fails most questions alone can be nearly as good a collaborator. Because human collaborators actively elicit reasoning, verify outputs, and extract value even on items the AI fails alone, standalone leaderboards dramatically overstate how much raw model capability translates into team performance. This would enable fine-grained comparisons of models' capacity to enhance the collective intelligence of human-AI teams in realistic problem-solving settings. Crucially, these contributions are interdependent: the ToM finding is only observable because the framework separates individual from collaborative ability---a distinction invisible to comparisons across means. Our work transforms human-AI collaboration from an anecdotal observation into a measurable, optimizable dimension of AI capability, enabling the design of systems explicitly aimed at emergent collaborative synergy rather than standalone performance.

\paragraph{Limitations and Future Work.} While we validate our framework using well-structured academic tasks (adapted from MMLU), these items represent a small subset of real-world collaborative contexts. Because items are multiple-choice with a performance ceiling, part of the equalizing effect reflects limited headroom for high-ability users. Our data also include only two models with users nested within them; while the random assignment of users to models identifies the aggregate synergy benchmark, the nested design prevents us from exploring individual-level heterogeneity in compatibility. 
Future work should examine whether our approach generalizes to other collaborative settings such as creative group work, workflow planning, or software development. 
We note that while the present study focuses on individual human-AI dyads, our framework naturally generalizes to multi-agent configurations. Additional hierarchical terms could model group-level collaborative ability in human teams or capture differential contributions from multiple AI assistants. Such settings may exhibit important emergent phenomena that are not observable in dyadic interactions. Developing and validating these multi-agent extensions represents an important direction for understanding collective intelligence in hybrid human-AI systems.

\section*{Methods}\label{sec:method}

\paragraph{Study Design and Data.}\label{sec:study_design}
Our data come from the ChatBench assessment constructed by \citet{Chang2025Chatbench}. ChatBench adapts MMLU \citep{Hendrycks2021MMLU} into a human-AI benchmark with three components: i) solo AI; ii) solo human; iii) human-AI pairs. ChatBench contains multiple-choice questions on three domains: math, physics, and moral reasoning. The dataset includes 396 questions covering a wide range of difficulties. Questions were randomly assigned to users and shared across solo and joint phases, ensuring direct comparability. On average, each question was answered by around 20 different users---about 5 users in solo work and 15 jointly with AI---providing overlap that supports stable item-level estimates.

Overall, the data contains responses from 667 humans. Participants begin the study by answering three questions by themselves (solo human). This enables independent estimation of individual ability \citep{Weidmann2021}. 
In Phase 2, users receive AI assistance via a chat interface and are randomly assigned to either \gpt\ or \llama\ (fixed at the user level) and answer nine more questions. 
For more details, see \citet{Chang2025Chatbench}.\footnote{In Phase 2, participants were also randomly assigned to two different modes of AI interaction: ``answer first'' in which humans provided a provisional answer before gaining access to AI assistance, and ``direct-to-AI'' in which participants had immediate access to AI help. In our analysis, we either use only data from the direct-to-AI condition or control for these different conditions using fixed effects. Note also that users had a direct piece-rate financial performance incentive to answer questions correctly (\$0.10 per correct answer).} 

\paragraph{Decomposing AI Impact Across- and Within-Users.}
We analyze the data using a form of multilevel modeling known as Item Response Theory \citep[IRT,][]{baker2004item} and a Bayesian workflow \citep{gelman2020bayesian}. Hierarchical Bayesian and IRT frameworks are increasingly used to evaluate AI systems themselves \citep{luettgau2025hibayes,zhou2026lost}, typically estimating a single ability per model. We instead treat the human–AI pair as the unit of analysis, estimate two distinct abilities for each person, and develop a bilinear extension anchored through experimental design features. Our data have three structural components: questions (items), users, condition (solo vs.~joint with AI vs.~AI alone). Our framework allows us to decompose the binary response (of answering questions correctly) along two crucial dimensions using random intercepts and slopes: $\Pr(Y_{ij} = 1) = \operatorname{logit}^{-1}(\text{Ability}_i-\text{Difficulty}_j)$, where Ability$_i$ represents the problem-solving capacity of user $i$ (either working solo or jointly with AI) and Difficulty$_j$ represents the difficulty of question $j$. The item-level effects $\beta_j$ and $\gamma_j$ are sufficient statistics for all stable item-level heterogeneity---domain, MMLU subject, format, answer-choice ambiguity, and AI standalone accuracy---so these sources cannot masquerade as user-level collaborative ability.

Using this structure, we estimate the problem-solving capacity of users working alone ($\theta$) and working jointly with AI ($\kappa$):
\begin{equation}
\Pr(Y_{ij} = 1) =
\begin{cases}
\operatorname{logit}^{-1}(\theta_i^{\text{human}} - \beta_j), & \text{if } joint_{ij} = 0 \; (\text{solo})
\\[6pt]
\operatorname{logit}^{-1}\!\Big(
\smash{\underbrace{\kappa_{i,\text{AI}}^{\text{total}}}_{\text{Ability}}}
-
\smash{\underbrace{(\beta_j + \gamma_j)}_{\text{Difficulty}}}
\Big) & \text{if } joint_{ij} = 1 \; (\text{with AI}).
\end{cases}
\label{eq:model1}
\end{equation}

Here, $\theta_i^{\text{human}}$ is user $i$'s latent ability when working alone, estimated from responses in the solo condition. $\kappa_{i, \text{AI}}^{\text{total}}$ represents the overall performance of user $i$ working with AI. Question difficulty is decomposed into solo difficulty ($\beta_j$), and difficulty when working collaboratively with AI ($\gamma_j$). Since the same questions appear in both solo and joint conditions, $\theta$ and $\beta$ are identified by solo performance, and joint parameters ($\kappa$, $\gamma$) are strictly identified by within-user, within-item variation.




This model allows us to estimate the \emph{boost} that each individual receives from working with AI. We define \emph{boost} as 
$\kappa_{i, \text{AI}}^{\text{total}} - \theta_i^{\text{human}}$ (estimates based on Eq.~\ref{eq:model1}). 
Conceptually, \textit{AI boost} reflects emergent synergy---gains arising from interaction, clarification, scaffolding, or co-construction. While this includes the AI's direct contributions---the AI provides the correct answer and the user adopts it---it depends on the user writing a useful prompt and deciding to adopt the AI answer. The boost should therefore be interpreted as a behavioral measure of the emergent human-AI  performance, including the full range of mechanisms through which human-AI collaboration affects performance. We discuss how our modeling framework naturally generalizes to multi-model and crossed designs in \supp~\ref{sec:generalize}. 

To benchmark the contribution of different AI models, we can further decompose joint performance into the collaborative ability of the user ($\kappa_i^{\text{human}}$) and estimates for different AI models via fixed effects ($\kappa_m^{\text{AI}}$; see Eq.~\ref{eq:model2} in \supp). Here, we can interpret $\kappa_{i}^{\text{human}}$ as the collaborative ability of user $i$ (when working jointly with AI). This collaborative ability includes the user's ability in deciding when and how to delegate tasks to the AI, how to formulate the delegation (i.e., writing LLM prompts), deciding whether to accept the AI's response, requesting clarification, and so forth. Similarly, $\kappa_{m}^{\text{AI}}$ represents the collaborative capability of the AI model (that user $i$ is paired with). This approach is an important extension of AI benchmarks as it quantifies the extent to which different models amplify human performance. In our validation data each user is paired with a single model, so the model contrast is identified between users and recovers the average causal effect of model assignment (users are randomized to models; \supp\ \ref{sec:generalize}). 

Finally, to test whether collaborative performance merely reflects existing domain expertise or draws on a distinct skill dimension, we relax the independence assumption in the priors and model the joint covariance structure of ($\theta$, $\kappa$), capturing both variance compression (equalization) and rank reshuffling ($\rho<1$). To test how AI combines with each skill---distinguishing substitution from multiplicative complementarity---we extend this framework into a three-arm bilinear model \citep{hoff2005bilinear} that anchors item-level human difficulty ($d_j$) and standalone AI capability ($\eta_{jm}$) using solo human and AI benchmark trials, and interacts both item attributes with $\theta_i$ and solo-orthogonal collaboration skill ($\kappa_i^\perp \perp \theta_i$; see Eq.~\ref{eq:2x2} in \supp~\ref{si:mechanism}).

%


\paragraph{Decomposing Linear and Non-Linear Synergy.} \label{sec:decomposition}
A key critique of static benchmarks is their inability to distinguish generalizable improvements from idiosyncratic interactions. To distinguish whether AI acts as a truly adaptive partner (non-linear synergy with specific user-item ``chemistry'') or merely a generally strong tool (additive gains), we perform several tests to distinguish linear from non-linear synergy. First, we test for non-additive synergy directly by asking whether the AI boost bends
with item difficulty in a user-specific way. We complement this with a simulation-based power analysis to bound the strength of non-linearity our analysis is able to detect (\supp~\ref{sec:power}). Second, we estimated a 2PL extension of the model that allows item-specific discrimination parameters. Finally, we propose a dispersion test in which we decompose human-AI synergy into a systematic and idiosyncratic factor:
\begin{equation}
\label{eq:synergy_decomp}
\operatorname{logit} \Pr(Y_{ij} = 1) = 
\underbrace{ (  \kappa_{i,\text{AI}}^{\text{total}} - (\beta_j + \gamma_j)  ) }_{\substack{\text{Additive} \\ \text{Synergy}}} 
~~+ 
\underbrace{\epsilon_{ij}}_{\substack{\text{Non-Linear} \\ \text{Synergy}}}
\end{equation}
where the first term represents \textbf{Additive Synergy} (stable user ability $+$ item difficulty shifts), and $\epsilon_{ij}$ represents \textbf{Non-Linear Synergy} (transient user-item dynamics). For example, one user might benefit from AI assistance on a physics problem but gain little on a moral reasoning question, while another user shows the opposite pattern. By fitting the additive baseline (Eq.~\ref{eq:model1}) and analyzing the dispersion of the residuals $\epsilon_{ij}$, our framework offers a novel diagnostic: it quantifies the extent to which collaboration is stable (trait-based) vs.~dynamic and non-linear (interaction-based). If dispersion of $\epsilon_{ij}$ is $\approx 1.0$, synergy is stable and trait-based. If it is $\gg 1.0$, synergy is context-dependent and non-linear. As an additional test, the random-slopes of our model capture heterogeneous effects across ability and task difficulty, allowing decomposition of human-AI synergy beyond the uniform gains assumption of a simple additive model.

\paragraph{Estimation.} Models are fit via the \texttt{brms} package in \texttt{R} \citep{burkner2017brms}, with weakly informative priors (8 chains and 8,000 iterations) as a default. Partial pooling regularizes estimates and improves generalization with limited or uneven data \citep{gelman2007data}. Although each user completes only three solo items, our fully Bayesian framework stabilizes these estimates via partial pooling and propagates remaining uncertainty into all downstream posteriors. Convergence diagnostics confirm robustness (\supp~\ref{sec:robustness}) and prior sensitivity analyses confirm that even aggressive shrinkage leaves the structural conclusions unchanged (\supp~\ref{sec:priors}). All reported results are from converged models with $\hat{R}$ values in the commonly accepted range \citep{burkner2017brms}. 



\section*{Contribution statement}
C.R.~and B.W.~conceived the study and developed the theoretical analysis. C.R.~conducted the computational experiments and analyzed the data. All authors wrote and reviewed the manuscript.

\section*{Acknowledgments and disclosure of funding}
The authors received no funding for this research. 
The authors declare no competing interests.

\section*{Ethics declaration}
This study is a secondary analysis of two publicly available datasets: the de-identified ChatBench dataset \citep{Chang2025Chatbench} and the data released by \citet{bansal2021does}; no new data were collected from study participants. 
Gold-standard ratings were produced as paid annotation work and collected no information about the annotators.

\section*{Data availability}
The ChatBench data are available at \url{https://huggingface.co/datasets/microsoft/ChatBench}. The data of \citet{bansal2021does} are available at \url{https://github.com/uw-hai/Complementary-Performance}.
Replication code will be available on \url{https://github.com/riedlc}.




\bibliography{references}
\bibliographystyle{icml2026}

\newpage
\appendix
\renewcommand{\thefigure}{A\arabic{figure}}
\setcounter{figure}{0}
\renewcommand{\theequation}{A\arabic{equation}}
\setcounter{equation}{0}

\onecolumn
\section*{Supplementary Information}

\section{Additional Equations for Bayesian Framework Estimation}
The main model introduced in the main text decomposes the contribution of the individual and the AI. We can further decompose the monolithic AI contribution into the contribution of different AI models. This approach can be used to \textit{benchmark} different models. To illustrate this, different AI models are indexed as $AI_m$:\footnote{The $m$ subscripts can also be applied to Eq.~\ref{eq:model1}, but we suppressed these for clarity of exposition.}

\begin{equation}
\Pr(Y_{ij} = 1) =
\begin{cases}
\operatorname{logit}^{-1}(\theta_i^{\text{human}} - \beta_j), & \text{if } joint_{ij} = 0 \; (\text{solo})
\\[6pt]
\operatorname{logit}^{-1}\!\Big(
\smash{\underbrace{\kappa_i^{\text{human}} + \kappa_m^{\text{AI}}}_{\text{Ability}}}
-
\smash{\underbrace{(\beta_j + \gamma_j)}_{\text{Difficulty}}}
\Big), & \text{if } joint_{ij} = 1 \; (\text{with } AI_m).
\end{cases}
\label{eq:model2}
\end{equation}
\vspace{0.5cm}

\paragraph{Generalizing to Multi-Model and Crossed Designs.} \label{sec:generalize}
While data in our validation study pairs users with only a single AI model, the proposed Bayesian framework is designed to scale to complex, crossed evaluation designs. In settings where users interact with multiple models, the framework can naturally be extended with crossed random effects ($\kappa_i^{\text{human}} + \kappa_m^{\text{AI}} + \delta_{i,m}$). This would allow researchers to benchmark model portfolios by mathematically decomposing performance into User Ability, Model Capability, and the User-Model Compatibility (interaction term; not identified here). We leverage this flexibility to explore the role of ToM by adding it as an interaction term in this way (see \supp~\ref{supp:ToM_method}). This flexibility makes the framework a general-purpose tool for disentangling the ``who,'' ``what,'' and ``how'' of human-AI synergy beyond static leaderboards.

\section{Additional IRT Modeling Results} \label{sec:robustness}
Our dataset contains three questions (items) per user in the solo condition. One concern may be that three items may be insufficient to precisely estimate user-level ability. However, our hierarchical model uses partial pooling to borrow strength across users and items. Regarding measurement precision, we observe a population standard deviation of $\sigma_\theta = 0.85$ and a median posterior SE of 0.91. In a standard fixed-effect framework, this noise-to-signal ratio could be problematic. However, our hierarchical Bayesian approach leverages partial pooling (shrinkage) to stabilize these estimates. While the reliability of any single user's rank is limited (as expected with k=3 items), the aggregate posterior distribution accurately recovers the population structure. MCMC diagnostics confirm excellent computational convergence ($\hat{R}$ = 1.00, Tail ESS > $7{,}000$), ensuring that the integration over this uncertainty is precise. Furthermore, the posterior estimates maintain a strong correlation with raw accuracy ($r=0.75$), validating that the latent trait recovery preserves the rank-order signal of the data despite the inherent uncertainty. Additional sensitivity tests with more aggressive shrinkage are robust as well, suggesting that noisy $\theta$ estimates from sparse measurement are not driving the result (\supp~\ref{sec:priors}).

To further assess the reliability of our solo-ability estimates, we leverage the subset of participants who completed two independent solo phases (six items total) and fit a bivariate latent model that estimates a separate solo ability for each phase, allowing us to recover the disattenuated correlation between them. The two latent abilities correlate at r = 0.82 (95\% CrI [0.61, 0.98]), indicating that individual solo ability is recovered consistently across independent item sets and is not an artifact of the three-item measurement. We also refit the headline results on the 6-item subsample and with propagated $\theta$ uncertainty; the $\theta \neq \kappa$ separation result is unchanged (an independent random-effects decomposition test recovers a solo-collaboration ability correlation well below one ($\rho = 0.64$ [95\% CrI: 0.48, 0.79]); nested model test $\Delta$ELPD = 21.0, SE = 7.4). In summary, the $\theta \neq \kappa$ separation persists when solo ability is measured with twice as many items, ruling out that it reflects noise in the three-item estimates. See also \supp~\ref{sec:priors} which investigates shrinkage as further evidence.

We also test whether \gpt\ or \llama\ is better at complementing users at different ability levels. For example, higher-ability users might benefit more from stronger models due to their higher capacity to evaluate and leverage suggestions. Conversely, lower-ability users might benefit more from stronger models due to better scaffolding and structured thinking support that such models might provide. We reject the hypothesis of heterogeneous effects: the impact of both models is stable across the distribution of solo ability ($\kappa_{\text{Llama}}^{\text{AI}}$ 0.02 [95\% CrI: -0.12, 0.17] compared to $\kappa_{\text{GPT4o}}^{\text{AI}}$ baseline).

\section{Background on Theory of Mind in Human-AI Settings} \label{sec:ToM_literature}
To better explain AI's impact, we draw on established theories from human-human collaboration, particularly Theory of Mind (ToM). ToM refers to the capacity to represent and reason about others' mental states \citep{premack1978does}. It plays a crucial role in human-human interaction \citep{nickerson1999hidden,lewis2003measuring}, allowing individuals to anticipate actions, disambiguate and repair communication, and coordinate contributions during joint tasks \citep{frith2006neural,clark1996using,sebanz2006joint,tomasello2010origins}. ToM has repeatedly been shown to predict collaborative success in human teams \citep{Weidmann2021,woolley2010evidence,riedl2021quantifying}. 
ToM is the glue that binds autonomous agents into a collective epistemic agent \citep{clark1998extended,tollefsen2006extended}. 
Its importance is also recognized in AI and LLM research \citep{prakash2025language,liu2025mind}, for purposes such as inferring missing knowledge \citep{bortoletto2024limits}, aligning common ground \citep{qiu2024minddial}, and cognitive modeling \citep{westby2023collective}. 
We argue that working with LLMs similarly requires adaptive, interaction-sensitive behavior due to the dialogic nature of human-AI interactions and the semi-autonomous, unpredictable nature of LLM output.\footnote{Early empirical work found that users benefit from forming accurate mental models of AI \citep{bansal2019updates,paleja2021utility,bansal2019beyond,gero2020mental,alipour2021improving}, though this effect is not always consistent \citep{fugener2022cognitive}.}

When teaming with generative AI like LLMs, users similarly need to anticipate their behavior and adapt accordingly given their semi-autonomous and unpredictable nature. Such adaptation is necessary as LLMs appear to users as having some level of agency and unpredictability. Four aspects of contemporary generative AI systems stand out. 
First, they possess meaningful agency by generating novel content without explicit commands, creating unpredictability that requires coordination \citep{dong2025humanizing}. Second, successful collaboration with LLM systems depends on the same dialogic mechanisms as human teamwork: eliciting outputs and providing feedback based on an evolving understanding of the partner's capabilities and goals \citep{clark1991grounding}. Third, deciding whether to accept, reject, or further interrogate LLM responses depends on estimating the LLM's competence and ``beliefs''. Fourth, LLMs exhibit behavior that strongly invites attribution of intentional states, encouraging users to ascribe beliefs \citep{gopnik1992child}. This tendency toward anthropomorphism is not just a cognitive bias but reflects the genuinely interactive nature of AI responses.\footnote{The human tendency to treat nonhuman agents as humanlike and apply social heuristics and mental-state inferences to interactive systems was documented even with pre-AI software \citep{reeves1996media,nass2000machines,weizenbaum1966eliza,waytz2010social}. Whether LLMs have a generalizable ToM is currently an open question, see, for example, \citet{prakash2025language}.}

Robust coordination with such agents benefits from users' ability to form and revise mental models of the AI's knowledge and capabilities. Misaligned expectations about the AI's capabilities, knowledge, or ``goals'' are likely to hinder task performance. That is, successful AI partnerships require users to engage in dialogue with their AI partner, form accurate mental models of what the AI knows and doesn't know, and provide clear instructions informed by this understanding \citep[cf.][for similar arguments]{laban2025llms}. When the AI makes errors, effective collaborators troubleshoot by hypothesizing what information the system lacked and updating their mental model accordingly.\footnote{This framework stands in contrast to traditional tool use, such as working with a notebook \citep{clark1998extended}. A notebook functions as a pure storage device—it cannot act independently, making its ``behavior'' entirely predictable. Effective notebook use relies on meta-cognitive awareness of one's own memory limitations rather than theory of mind about another agent's capabilities. As a result, individuals skilled at self-monitoring and organizing information benefit most from notebooks, not those adept at understanding and coordinating with other minds. Working with a notebook, users are not required to form a mental model of why the notebook inserts certain keywords, whether it will auto-correct technical terms, or when it might reorganize pages.} In human-AI collaboration, ToM may enable users to form accurate mental models of the AI's capabilities and limitations, creating effective coordination (despite the AI's lack of mutual intentionality). ToM behavior may also support role differentiation, allowing users to leverage the AI's strengths, a division of cognitive labor resembling a transactive memory system in human groups \citep{lewis2003measuring,wegner1987transactive}.

Adopting a team-based perspective on human-AI interactions yields theoretical predictions which can be evaluated with human-AI benchmarks. First, we hypothesize that performance varies across conditions in a way that reflects two distinct, condition-specific latent abilities: individual problem-solving ability when working alone and collaborative ability when working jointly with AI. Evidence from human teams suggests that these two dimensions of human ability are separately identifiable, with each broad ability contributing independently to team success  \citep{Weidmann2021}. Second, given the importance of ToM in enabling effective coordination and adaptive interaction in human groups, we hypothesize that higher ToM capabilities will positively predict collaborative ability, but not solo ability when working alone. Third, we ask whether ToM operates as a direct channel for synergy in human-AI teams (while noting that our design can establish association rather than causation). Concretely, we test the idea that users with strong ToM provide prompts that lead AI systems to engage in higher-quality dialogue. Last, we build on recent advances in psychology suggesting that ToM is a dynamic construct \citep{Schneider2012Cognitive}. That is, ToM can be thought of as a stable individual trait whose expression is shaped by contextual affordances and cognitive demands. As such, we should expect ToM expression to vary within individuals \citep{qureshi2010executive,Lin2010Reflexively,meijering2010facilitative} and hypothesize that within-user dynamic \textit{increases} in ToM expression (i.e., when users exhibit unusually strong ToM) will be positively associated with the quality of AI responses. 

\section{Method Details for Theory of Mind Analyses}\label{supp:ToM_method}
We build on the decomposition in Eq.~\ref{eq:model2} to investigate a potential cognitive mechanism underlying effective collaboration between humans and AI. In principle, any user-level characteristic can be used to explore why some users have stronger collaborative abilities when working with AI. In Eq.~\ref{eq:model3}, we focus on Theory of Mind. As discussed in \supp~\ref{sec:ToM_literature}, evidence from the literature on collective intelligence and human teams suggests that ToM is a crucial element of successful teamwork.

\begin{equation}
\Pr(Y_{ij} = 1) = \begin{cases}
\operatorname{logit}^{-1}(\theta_i^{\text{Solo}} \times \mathrm{ToM}_i - \beta_j) & \text{if } joint_{ij} = 0 \text{ (solo)} \\
\operatorname{logit}^{-1}(\smash{\underbrace{\kappa_{i}^{\text{Team}} \times \mathrm{ToM}_i+\kappa_{\text{AI}[i]}^{\text{Team}}}_{\text{Ability}}} - \smash{\underbrace{(\beta_j + \gamma_j)}_{\text{Difficulty}}}) & \text{if } joint_{ij} = 1 \text{ (with AI)}
\end{cases}
\label{eq:model3}
\end{equation}
\vspace{0.5cm}

To measure ToM, we analyze each user's written prompts for signatures of Theory of Mind---the capacity to reason about others’ knowledge, beliefs, and intentions (see \supp~\ref{sec:ToM_LMRA} for a description of how ToM was measured). By integrating behavioral performance with linguistic indicators of ToM, we aim to better understand how users' mentalizing abilities contribute to the emergent human-AI collaboration synergy.

Last, we explore whether user ToM is associated with the quality of AI responses. First, we collect additional data on the AI response quality (see \supp~\ref{sec:AI_Quality} for details). We then exploit the panel data structure with repeated observations to perform a variance decomposition \citep{mundlak1978pooling}. This analysis decomposes each user's ToM signatures into a user-level mean, which captures trait-like differences across users, and a within-user deviation term, which reflects within-user, dynamic changes in ToM expression. Both elements could, in principle, relate to AI response quality.

We compute the user-level mean as a leave-one-out measure \citep{jacob2008can}. This cross-validation style estimation of the user-level trait helps to address concerns of overfitting and avoids contamination of the latent trait estimate. Using question-level $\text{ToM}_{ij}$ measures for question $j$ by user $i$ we compute

\begin{equation}
\overline{\text{ToM}}_{i(-j)} = \frac{1}{J_i - 1} \sum_{j' \ne j} \text{ToM}_{ij'}
\end{equation}

\begin{equation}
\text{ToM\_dev}_{ij} = \text{ToM}_{ij} - \overline{\text{ToM}}_{i(-j)}
\end{equation}

The ToM analyses include additional control variables for effort and proxies for prompt quality, all computed on the question level.
\paragraph{Turns.} A count of the number of back-and-forth conversation turns between the user and the AI. More turns indicate more effort, such as asking for clarification or providing refinements.
\paragraph{Total Characters.} A count of the number of characters written by the user across all conversation turns of the question. Writing longer prompts (controlling for the number of prompts) indicates more effort.
\paragraph{Edit Ratio.} We compute the edit ratio as the normalized Levenshtein distance between the user’s concatenated LLM prompt(s) and the original question, capturing the proportion of characters that were edited and thus reflecting the extent of user-initiated modifications. User prompts that simply copied and pasted the original question have an edit ratio of 0 and would thus indicate low effort.
\paragraph{Concreteness.} As a lexical proxy for prompt specificity, we measure the mean concreteness of the words in each prompt using the norms of \citet{brysbaert2014concreteness}, who provide human concreteness ratings (on a 1--5 scale, from abstract to concrete) for roughly 40{,}000 English lemmas. We tokenize the human-written prompts, match tokens to the norm list, and average the ratings over matched words.
\paragraph{Writing Complexity.} We measure the readability and complexity of the prompt via the ``Simple Measure of Gobbledygook'' (SMOG) which captures polysyllabic word density \citep{mcLaughlin1969smog}.
\paragraph{Spelling Mistakes.} Spelling mistakes introduced by sloppy prompt writing are one obvious proxy for prompt quality. We measure spelling mistakes---entered as the total count of mistakes---using the Hunspell database \url{https://hunspell.github.io/}.

\section{Theory of Mind Measurement} \label{sec:ToM_LMRA}

We use LLMs to perform text analysis of users' written prompts. Prior work has shown that modern LLMs can do this with high accuracy and reliability, achieving scores above those of trained human coders \citep{rathje2024gpt}. It is sometimes referred to as a Language Model as a  Research Assistant \citep[LMRA;][]{eloundou2024first}. We used the LLM prompt below and \verb|gpt-4.1-2025-04-14| with temperature set to 0. The assessment is performed on the question level, concatenating all user prompts (if there were multiple; but excluding the AI responses). The analysis builds on ideas of using computational text analysis indexing dynamic collective constructs from observed interaction text \citep{mathieu2022indexing}. Prior work has established the validity of analysis of recorded conversations to assess theory of mind capacity, especially in the context of collaboration success \citep{hennessy2016developing,knight2017collaborative,kim2021theory,alkire2023theory} and the relationship between traditional ToM measures such as performance on false-belief tests, competence at reading others' minds in everyday conversational interactions, and other ToM measures like the Reading the Mind in the Eyes Test \citep{de2014talking}. The prompt design relies on chain-of-thought reasoning with in-context examples \citep{wei2022chain}. The specific markers of ToM are compiled from prior work on ToM, in particular recent reviews of ToM in psychology \citep{beaudoin2020systematic,fu2023systematic,quesque2020theory}, work linking ToM to rich signals in communication such as clarification requests and communication repair \citep{tomasello2003makes}, work in computer science \citep[e.g.,][]{bara2021mindcraft}, and older ideas around social intelligence \citep{kihlstrom2000social}.

As a robustness test and to address concerns of style confounds or model family effects, we also collected Theory of Mind assessments using Google's \verb|Gemini 2.5 Pro|---an entirely different family of model, trained by a different research lab---and find substantially the same results (see Footnote \ref{fn:ToM_robustness}). 

To complement this main analysis, we also collected data on a narrower interpretation of Theory of Mind, restricting the operationalization to mental-state reasoning (belief tracking, knowledge gap detection, epistemic monitoring, capability attribution) and explicitly excluding broader aspects (such as establishing rapport; see full prompt in our GitHub repository). 
We again used \verb|gpt-4.1-2025-04-14| with temperature set to 0. As expected, this narrower measure shows substantially lower variation across users (SD = 0.37 compared to 0.94 for the broader prompt). However, the present design does not directly establish ToM as a primary channel, and we note that substantial residual heterogeneity in individual AI benefits (random slope SD = 1.05) indicates that ToM does not account for all individual-level variation.

\begin{tcolorbox}[llmprompt, title=ToM Signatures in User Prompts]
\setlength{\leftmargini}{12pt} 

Context: You are an expert research assistant. 

Evaluation Task: You are evaluating the degree to which the dialogue by a human user with an AI assistant show signs of Theory of Mind (ToM) of the human user. ToM refers to the ability to attribute mental states—such as beliefs, intents, desires, or knowledge—to oneself and others. Evaluate questions on a scale from 0 (no sign of theory of mind) to 5 (multiple signs of theory of mind present). In addition to the question asked by the user (USER DIALOGUE), you are also provided the original question (ORIGINAL QUESTION) that the user is trying to answer with the help of AI so that you can assess if and how the user has modified their query.

In your evaluation, consider the following signs of theory of mind.

Indicators of high theory of mind:
\begin{itemize}
    \item Establish rapport: e.g., by introducing themselves (``hello''), by acknowledging responses (such as ``thanks'' and ``okay'').
    \item Belief tracking: e.g., inference of AI's beliefs about world state and tracking how they may change dynamically during conversation.
    \item Knowledge gap detection: Identifying information asymmetries between the user and the AI.
    \item Perspective-taking: Understanding AI's visual and spatial reasoning, and that it may differ from the user's own. Awareness about what the AI needs to know (vs.~what is irrelevant) to work on the problem together.
    \item Goal inference: Predicting the AI assistant's intended goals from dialogue.
    \item Communication intent: Reasoning about why specific information was shared or what the users goals are (e.g., ``Help me solve this questions'', ``I already solved the problem'').
    \item Mental state attribution: Understanding that the AI as unique knowledge, capabilities, and limitations. 
    \item Plan coordination markers: Linguistic signals of strategy alignment and developing a collaboration plan (e.g., ``I'm going to ask you a question'').
    \item Explanatory dialogue: Information provision addressing knowledge gaps such as providing context about their own knowledge level (e.g., ``I'm a beginner in physics''), setting expectations (e.g., ``I need a comprehensive explanation''), and goals (e.g., ``Explain this like I'm 15'').
    \item Confirmation seeking: Verification behaviors indicating uncertainty, including asking for explanations and clarification (e.g., ``Is my approach correct?'').
    \item Positioning and coordination: Awareness and coordination of different viewpoints
    \item Build on ideas: Understanding of others' mental states with elaboration (e.g., ``this is not what I meant'' and ``the answer should be formatted as a fraction'').
    \item Reference back: Explicit acknowledgment showing memory of partners' thoughts
    \item Challenge/disagree: Sophisticated engagement demonstrating ToM understanding such as pointing out mistakes (e.g., ``could you be wrong?'').
    \item Epistemic markers: Language revealing understanding of knowledge states (``I think you know...'')
    \item Justification requests: Questions showing awareness of reasoning processes
    \item Evaluative language: Understanding of partners' credibility assessments
    \item Metacognitive references: Explicit discussion of thinking processes (e.g., ``right, that is a difficult problem'').
    \item Conversational adaptation: Real-time adjustment to partner's communication style.
\end{itemize}

Absence of the above ``high theory of mind'' markers should be classified as low levels of theory of mind. Additionally, consider these indicators of low theory of mind:

\begin{itemize}
    \item Sharing irrelevant information: failing to detect relevant knowledge asymmetries. 
    \item Assuming shared context without previously establishing it (e.g., saying ``answer this question'' without specifying what the question actually is).
    \item Delegating a trivial question (e.g., ``how many years are in a decade'') or moral questions for which the human user clearly has the advantage (e.g., ``is peeing in a pool bad'') are signs of low theory of mind.
    \item Misunderstanding the AI's unique knowledge, capabilities, and limitations such as treating it as Google (e.g., ``stomach hurts sharp pain'' indicates the user is search for a website rather than asking a specific question that is more aligned with the AI's capabilities).
\end{itemize}

Instructions: 

1. Analyze whether the question shows signs of theory of mind thinking. 

2. Clearly explain your reasoning in a paragraph.

3. Then, in a separate line, output the classification in the following format:

FINAL ANSWER: [0-5]
\end{tcolorbox}

\section{Validation using Human Ratings} \label{sec:Human_gold}
To validate the approach of studying dialogue-level ToM signatures in human-AI collaboration and the LMRA assessment, we collected gold standard human ratings for a stratified (by ToM level) random sample of 120 dialogues. We recruited 32 subjects from Prolific (general US pool), each of whom rated 19 dialogues on average (SD=14) yielding an average of five ratings per dialogue. We paid human raters \$1.50 for reading the instructions, \$0.25 per rated dialogue, and a \$1.50 bonus if they rated more than 10 dialogues (to encourage raters to evaluate several dialogues to increase consistency). The rating instructions were mostly identical to those used for the LLM with minor changes to formatting (e.g., breaking instructions up into several pages) and the addition of information about compensation.

Individual human raters showed low reliability (ICC(2) = 0.13), reflecting some inherent noise in single ratings. The aggregated gold standard showed excellent reliability (ICC(2,k) = 0.83 [95\% CI: 0.78, 0.87]). The high inter-rater reliability among independent human judges demonstrates that dialogue-level ToM signatures in human-AI collaboration can be assessed consistently, validating the construct and supporting the use of this evaluation approach.
The LMRA's ratings demonstrate substantial agreement with the gold standard (ICC(2) = 0.42; ICC(2,k) = 0.59 [95\% CI: 0.42, 0.72]), outperforming a single human rater and approaching the consistency of the aggregated human consensus. 
Because single human ratings are themselves noisy (ICC(2) = 0.13), agreement with the gold standard is bounded from above by noise in the criterion; the LMRA-gold correlation should therefore be read as a lower bound on validity.
All ICC estimates between the LMRA and the human gold standard are significant ($p < 0.001$). These results support the use of an LMRA as a proxy for human assessment, and demonstrate the promise of LLM-based assessments as scalable approximations of human judgment for dialogue-level ToM. However, reliability of individual human ratings is low and the findings may be task-specific. 

\section{Construct Validity of LMRA ToM Ratings} \label{sec:LMRA_ToM_construct}
\begin{figure}[h!]
\begin{center}
\includegraphics[width=.6\textwidth]{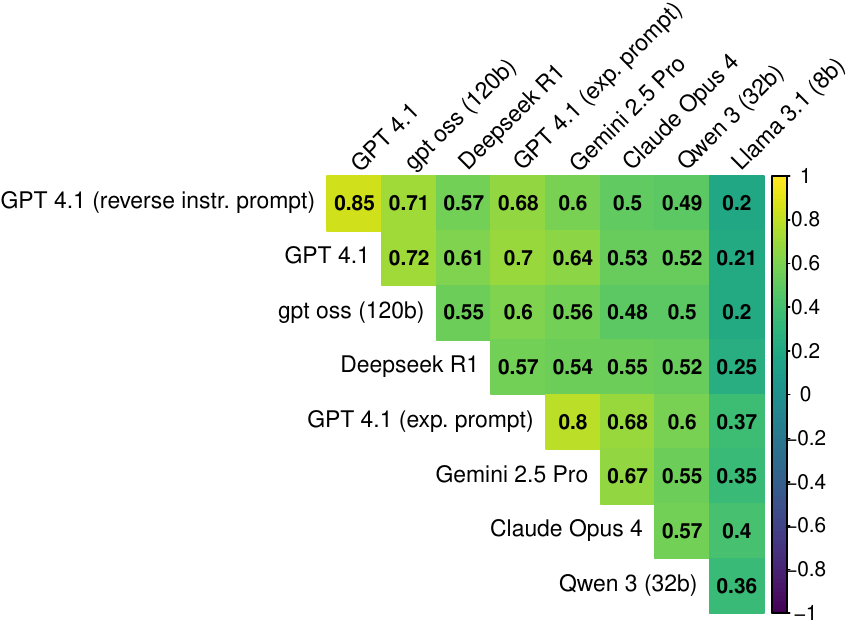}
\end{center}
\caption{
    \textbf{Pairwise inter-coder reliability (ICC(2))) of LMRA ToM assessment.}
}
\label{fig:ToMICC}
\end{figure}

To further establish the construct validity of LMRA ToM ratings of user dialogues we took several additional steps to implement emerging LLM-as-a-judge best practices \citep{li2025generation}. First, we inspected dialogues for signs of adversarial prompt injection \citep{maloyan2025adversarial} but found no signs of it. This is not surprising given that users were  unaware that we would code their dialogues for signs of ToM and that they were incentivized to achieve high question answering performance. 
Next, to address concerns of style confounds or model family effects, we replicated data collection using six other models: Google's \verb|Gemini 2.5 Pro|, Anthropic's \verb|Claude Opus 4|, OpenAI's \verb|gpt-oss-120b|, Alibaba's \verb|Qwen3 32B|, DeepSeek's \verb|Deepseek-R1|, and Meta's \verb|Llama-3.1-8B|. All models were accessed via the OpenRouter API. To assess robustness to prompt bias, we also collected data using an expanded ToM‑assessment instruction set and alternative rating schemes (point-based assessment vs.~pairwise comparisons). To explore ordering bias, we collected data using a prompt which reverses the instructions for high vs.~low ToM signatures. Across all nine LMRA ``raters'' we find excellent \citep{cicchetti1994guidelines} ensemble inter-coder reliability (ICC(2,k) = 0.90 [95\% CI: 0.85, 0.92]). We also show all pairwise ICC(2) scores (Figure \ref{fig:ToMICC}). The smaller \verb|Llama-3.1-8b| shows substantially lower agreement with the other models, consistent with limited capacity to judge ToM reliably. As a sensitivity analysis, we exclude this model which increases ICC(2,k)) to 0.92 [95\% CI: 0.91, 0.94]. Taken together, these findings provide convergent evidence that LMRA ToM ratings validly capture the intended construct, are robust across models and prompts, and suggest that any single model's ratings are representative of the ToM construct.

Finally, to test robustness to judgment bias we collected data using a prompt that asked the LMRA to make pairwise comparisons as to which dialogue shows a higher degree of ToM (instead of asking for assessment on 0-5 rating scale). We collected data on 12{,}000 pairwise comparisons drawing two random dialogues using \verb|gpt-oss-120b|. To limit order bias, options were given neutral labels without a natural order. Next, we consider these pairwise judgments ``matches'' between the users who wrote them and aggregate them in an Elo-based ranking (ranking all users in the study by their ToM prompt writing capacity; based on about 37 pairwise judgments per user; min = 1164, mean = 1500, max = 1793). We find moderate-to-strong rank association between the baseline \verb|gpt-4.1| average rating and the Elo-based ranking ($\rho_s = 0.61$; $p < 2.2 \times 10^{-16}$). After calibrating the Elo scores to the 0-5 rating scale (via linear regression), we find moderate-to-good absolute agreement ICC(3,1) = 0.68 [95\% CI: 0.64, 0.72] with \verb|gpt-4.1| scores and ICC(3,1) = 0.77 [95\% CI: 0.73, 0.80] with \verb|gpt-oss-120b|. These results indicate that the ToM construct can be reliably assessed using both absolute and relative judgment prompts.

\section{AI Response Quality Measurement}\label{sec:AI_Quality}
To assess the quality of AI responses we used \verb|gpt-4.1-2025-04-14| with temperature set to 0 and the prompt below. 
The assessment was done on the question level, concatenating all LLM responses if there were multiple turns of interaction (but excluding the user's prompts).

\begin{tcolorbox}[llmprompt, title=AI Response Quality]
Context: You are an expert research assistant. A user is working with an AI assistant to answer a question (see ORIGINAL QUESTION below). The AI gave the below response (AI RESPONSE).

Evaluation Task: You are evaluating the quality of the AI response given to the user. Evaluate AI responses on a scale from 0 (low quality, not helpful to answer the question) to 5 (high quality, very helpful to answer the question).
Consider not only whether the answer itself has the correct answer but how helpful the answer is to the user to correctly answer the question (e.g., by providing explanation and reasoning).

Instructions: 

1. Analyze the quality of the AI response. 

2. Clearly explain your reasoning in a paragraph.

3. Then, in a separate line, output the quality assessment in the following format:

FINAL ANSWER: [0-5]

ORIGINAL QUESTION: \{...\}.

AI RESPONSE: \{...\}
\end{tcolorbox}

\section{Causal Effect of Working with AI} \label{sec:learning}
The experimental design places users in the solo condition first and the with-AI condition second. One concern resulting from this design is that this may confound working with AI with temporal learning or fatigue effects since working with AI is always observed second. As a robustness test we leverage another feature of the \citet{Chang2025Chatbench} dataset. Specifically, the design included another randomly assigned treatment condition in which some users worked on a set of questions alone for a second time, before finally working with AI in the last step. Instead of modeling only two phases (solo vs.~joint) we model three phases (solo vs.~solo after practicing vs.~joint with AI). This model explicitly controls for and models any learning/fatigue effects. It thus avoids the confound between the AI condition and ``AI is always observed second''. We find consistent results with a slight negative effect of the second phase indicating fatigue (rather than learning). This indicates that our main effects somewhat underestimated the true benefit of working with AI ($\rho_p = 0.58$, $p < 0.001$). This model causally identifies AI effects (from random assignment of users to AI models), and phase-based learning/fatigue effects (based on phase progression with random assignment to conditions). 

We also perform additional robustness tests to confirm stable estimates of individual ability using different diagnostic tests. First, MCMC diagnostics show excellent convergence ($\hat{R}$ = 1.00) and very high effective sample sizes (Tail ESS between 7{,}000--26{,}000 effective samples per parameter), confirming that posterior estimates of user-level ability and phase-specific slopes are stable and reliable, even though each user completed only a limited set of items per phase. Furthermore, posterior estimates correlate strongly with raw accuracy ($r = 0.75$), demonstrating that the hierarchical model produces precise, interpretable ability estimates.



\section{Power Analysis of Non-Additive Synergy}\label{sec:power}
The conclusion that human-AI synergy is predominantly additive (Section Non-Additive Synergy in the main text) rests on a null result: the additive model showed no residual non-linearity. On its own this cannot distinguish a genuinely additive effect from a non-linear one too small for the design to detect. The following analysis quantifies that distinction by establishing how large a non-linearity we could have detected. 

Specifically, we test for non-linear synergy by comparing the additive model against an expanded model that lets each user's collaborative boost vary across item-difficulty levels, and ask whether that added user-by-difficulty term explains reliable variance. We treat the fitted additive model as ground truth, and inject a user-by-difficulty-bin interaction of known variance into the joint-condition trials. We then simulate responses on the observed design, and refit both the additive and an expanded (interaction) model through the identical hierarchical estimator used in the main analysis ($S=200$ parametric bootstrap simulations per effect size level). The analysis sweeps across different magnitudes of injected non-linearity to yield a detection-power curve and a minimum detectable effect. Power is the detection rate across the 200 replicates.

We find the procedure is well calibrated (false-positive rate $\leq$ 2\% with no injected interaction) and reaches $\geq$ 80\% power at an interaction accounting for 16\% of the additive-synergy variance. These results are consistent using both 3- and 5-bin difficulty partitions. Because the estimator refits the full shrinkage pipeline on simulated data, this bound already incorporates any shrinkage-induced attenuation, so shrinkage cannot explain the non-detection of non-linearity in the observed data.

Framed in production-function terms, this test explores whether user-specific returns to AI assistance depend on how difficult tasks are for humans. That is, net of the average item-level boost, do certain users gain disproportionately where tasks are hardest or see their advantage fade when questions get hard (AI acts as a ``crutch'' that helps them only when it is least needed)? Finding this margin additive implies that each user's collaborative log-odds boost is roughly constant across levels of question difficulty.

\begin{figure}[t!]
\centering
\includegraphics[width=.5\columnwidth]{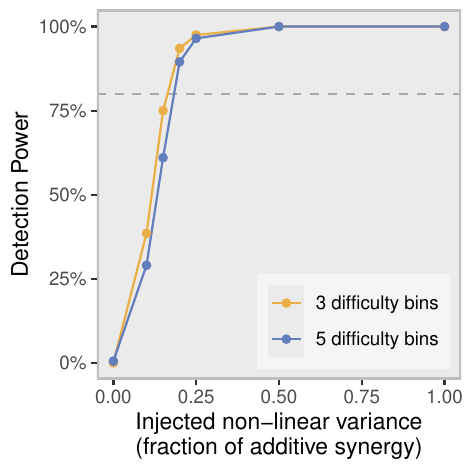}
\caption{
    \textbf{Power analysis to detect non-additive synergy in $\text{user} \times \text{difficulty-bin}$ terms.}
    Values based on parametric bootstrap simulation from the model based on Eq.~\ref{eq:model1} (with 200 simulations per effect size).
}
\label{fig:power}
\end{figure}

\section{Additional Insights into Mechanisms of Complementarity}\label{sec:complementarity}
To provide an additional window into the mechanism of complementarity we use model-based, item-level decomposition to explore item-level differences across difficulty and task domain. Specifically, we quantify shifts from answering incorrectly alone to answering correctly with the help of AI via posterior predictive counterfactuals from the fitted hierarchical IRT model (Eq.~\ref{eq:model2}). This has the added advantage that we can compute counterfactuals for each user-question for both models (\gpt~and \llama). 


On average, users switch from wrong answers to right answers on 25\% of the tasks when working with \llama~and 29\% when working with \gpt, indicating a substantial share of cases where collaboration moves responses from incorrect to correct. Users also switch in the opposite direction but this happens less often. Users switch from right-alone to wrong-with-AI 8.6\% with \gpt~and 12.3\% with \llama. Net switch rates are credibly positive for both models (\gpt: 20\% [95\% CrI: 17.4\%, 23\%]; \llama: 12.7\% [95\% CrI: 9.8\%, 15.5\%]).
Switch rates from wrong-to-correct answers increase monotonically with item difficulty for both models. On easy questions (1st decile) users switch 17\% (\llama) and 19\% (\gpt). On hard questions (10th decile) users switch 31\% (\llama) and 38\% (\gpt). Damaging switches similarly increase with question difficulty (from 7\% $\to$ 10\% with \gpt; 10\% $\to$ 14\% with \llama). This pattern is precisely what complementarity predicts: collaboration is most beneficial on harder items and in line with our results shown in Figure \ref{fig:IRT}A.

Next, we break out complementarity by user skill, cut into terciles (Table \ref{table:switch_rates}). Unconditionally, damage rises with user skill and gains fall. However, this gradient is fully explained by opportunity: damage requires being correct alone. Conditional on opportunity, per-opportunity gain and damage rates are statistically flat across ability terciles. Harm from AI collaboration is therefore not concentrated in any skill group; it scales with how much a user has to lose, at a uniform per-opportunity rate determined mainly by the AI arm (\gpt~helps more than \llama).

\begin{table}[!t]
\centering

\begin{tabular}{llc rrr rrr}
\toprule
 & & & \multicolumn{3}{c}{Unconditional} & \multicolumn{3}{c}{Conditional} \\
\cmidrule(lr){4-6} \cmidrule(lr){7-9}
Model & Direction & Skill Tercile & Mean & 2.5\% & 97.5\% & Mean & 2.5\% & 97.5\% \\
\midrule
GPT-4o & gain & 1 & 0.407 & 0.344 & 0.466 & 0.695 & 0.640 & 0.744 \\
 &  & 2 & 0.335 & 0.301 & 0.369 & 0.681 & 0.644 & 0.717 \\
 &  & 3 & 0.236 & 0.210 & 0.264 & 0.715 & 0.681 & 0.748 \\
 & damage & 1 & 0.062 & 0.040 & 0.086 & 0.149 & 0.105 & 0.196 \\
 &  & 2 & 0.080 & 0.064 & 0.096 & 0.157 & 0.129 & 0.185 \\
 &  & 3 & 0.095 & 0.080 & 0.111 & 0.142 & 0.120 & 0.163 \\
\midrule
Llama & gain & 1 & 0.354 & 0.296 & 0.413 & 0.605 & 0.546 & 0.661 \\
 &  & 2 & 0.288 & 0.255 & 0.320 & 0.587 & 0.545 & 0.630 \\
 &  & 3 & 0.206 & 0.181 & 0.231 & 0.624 & 0.586 & 0.661 \\
 & damage & 1 & 0.088 & 0.062 & 0.119 & 0.211 & 0.160 & 0.269 \\
 &  & 2 & 0.112 & 0.094 & 0.132 & 0.221 & 0.190 & 0.254 \\
 &  & 3 & 0.136 & 0.117 & 0.155 & 0.203 & 0.179 & 0.227 \\
\bottomrule
\end{tabular}
\vspace{1pt}

\caption{Switch rates based on model-based counterfactual simulation.} 
\label{table:switch_rates}
\end{table}

Decomposing the damage rate into exposure and susceptibility, $P(\text{harmful switch}) = P(\text{AI wrong}) \times P(\text{harmful switch} \mid \text{AI wrong}) + P(\text{AI right}) \times P(\text{harmful switch} \mid \text{AI right})$, shows that \llama's higher damage rate is driven almost entirely by exposure---its advice is wrong on 58\% of items versus 31\% for \gpt---rather than by differential credulity of its users (susceptibility of 12.9\% vs.~10.0\%). Because both counterfactual arms are evaluated on identical user--item pairs, this comparison holds user composition fixed; notably, users of the more accurate model are not more susceptible per error, contrary to the concern that greater trust induces greater over-reliance. The remaining harmful switches occur despite correct advice (5.6\% \gpt; 4.8\% \llama)---the majority of \gpt's total damage---indicating that as advice quality improves, residual harm increasingly reflects interaction friction rather than bad advice. No user (0 of 667) shows a credibly negative expected boost under either model, indicating that item-level damage is offset by gains at the level of every individual user.

Breaking out complementarity by domain we find the biggest complementarity for physics questions (32\% (\gpt) and 28\% (\llama)), followed by math (29\% (\gpt) and 25\% (\llama)), and moral reasoning (26\% and 22\%). The \llama~vs.~\gpt~difference is consistently about 4\% across the topic domains (this is in line with our model-based analyses where we do not find a significant interaction effect between models and item difficulty beyond the magnitude difference of \gpt~being the better model overall).

\section{Synergy, Not Substitution: Collaborative Gains Even When AI Is Wrong}\label{sec:substitution}
In this section we present additional analyses to explore whether there is genuine emergent synergy where the interaction produces something neither party would achieve alone---through scaffolding, clarification, co-construction, iterative refinement \citep{riedl2026cognitive,kelley2025personalized}. The alternative hypothesis is that the observed effect is the result of pure delegation and adoption: the AI just gives the right answer and the user adopts it. To sidestep the issue of the endogenously written user prompts (which may already contain signs of collaboration ability) we rely on the exogenous AI solo data. We add AI accuracy on any given question $j$ as a control to the model from Eq.~\ref{eq:model1}. The user-level SD for the joint slope drops from 1.56 to 1.14: a reduction of approximately 27\%. However, the estimate is still large and credibly different from zero [95\% CrI: 0.87, 1.41]. This suggests that AI solo ability does indeed absorb some of what looked like collaborative ability, but the majority of user-level variation in collaborative performance is not explained by simply being paired with an AI that already knew the answer. This estimate is conservative. The control absorbs all variance correlated with AI standalone accuracy, including the non-trivial skill of correctly calibrating reliance on AI output. Prior work demonstrates that users frequently over- or under-rely on AI even when its accuracy is known \citep{bansal2019beyond,fugener2022cognitive,Caplin2025ABC}, suggesting that appropriate adoption itself reflects meaningful collaborative ability. Furthermore, users need to write appropriate prompts for AI responses to include the correct answer, even if the AI is---in principle---able to answer the question correctly. These effects are naturally captured in our AI Collaboration Ability construct.

To sharpen this distinction further, we re-estimate the model from Eq.~\ref{eq:model1} splitting the joint condition into two indicators by whether the paired AI could solve the item on its own or not (arm-matched few-shot AI-alone accuracy above $0.5$, with correlated user-level random slopes on both). Pure delegation makes three predictions: users should not improve on items the AI cannot solve alone; user-level variance in collaborative performance should be concentrated on AI-solvable items (reflecting adoption propensity only); and performance in the two regimes should be only weakly related. We find the opposite on all three counts. We find users improve significantly over their solo baseline even on items the AI could not solve alone ($\beta = 0.80$ [95\% CI: 0.59, 1.01], compared to $\beta = 1.40$ [95\% CI: 1.16, 1.64] on AI-solvable items). User-level variation in the collaborative slopes is comparable across the two regimes (SD $= 1.08$ [0.75, 1.39] on AI-solvable vs.\ $0.88$ [0.52, 1.21] on AI-unsolvable items). And the two slopes are highly correlated at the user level ($\rho = 0.80$ [95\% CI: 0.53, 0.96]), indicating a unified interaction skill that operates regardless of whether the AI has the correct answer or not. Together with the finding that ToM predicts collaborative but not solo ability, we read this as evidence of an active, evaluative interaction process---scaffolding, error-checking, and elicitation---rather than passive adoption of AI output. This allows users to benefit from AI even in situations where the AI alone does not have the right answers.

\section{Sensitivity to Different Priors}\label{sec:priors}
To test if key estimates are stable or sensitive to specific priors, we conducted sensitivity analyses using different prior specifications (Table \ref{tab:priors}). We vary both the half-$t$ prior on random-effect standard deviations and the LKJ prior on the latent correlations of our main model (Eq. \ref{eq:model1}). Estimates are extremely stable: the collaboration boost SD is within $1.45$--$1.57$ and $\rho$ is virtually identical across all specifications, confirming that the structural conclusions are data-driven rather than dependent on priors. The stable $SD(\kappa - \theta)$ across prior specifications confirms that meaningful individual differences in collaboration ability persist regardless of how aggressively the model regularizes toward homogeneity. That is, even when priors actively shrink all users toward identical boost estimates, the data insist that people genuinely differ in how effectively they collaborate with AI.

\begin{table}[!h]
\centering
\centering
\begin{tabular}[t]{llll}
\toprule
Prior Specification & SD($\kappa - \theta$) & $\rho(\theta, \kappa - \theta)$ & Intercept\\
\midrule
Default (\texttt{brms}) & 1.55 [1.28, 1.84] & -0.89 [-0.93, -0.84] & 1.41 [1.23, 1.58]\\
Tighter SD & 1.52 [1.24, 1.80] & -0.89 [-0.93, -0.83] & 1.40 [1.22, 1.57]\\
Wider SD & 1.56 [1.28, 1.85] & -0.89 [-0.93, -0.84] & 1.40 [1.23, 1.58]\\
Shrink-to-zero corr & 1.49 [1.22, 1.77] & -0.87 [-0.92, -0.80] & 1.38 [1.20, 1.57]\\
Uniform corr & 1.57 [1.29, 1.85] & -0.89 [-0.93, -0.84] & 1.41 [1.24, 1.59]\\
\addlinespace
Tight SD + shrink & 1.45 [1.17, 1.73] & -0.87 [-0.91, -0.80] & 1.38 [1.20, 1.56]\\
\bottomrule
\vspace{1pt}
\end{tabular}
\caption{Sensitivity of key IRT parameters to prior specification.}
\label{tab:priors}
\end{table}

We repeat the prior sensitivity analysis with the model exploring the effect of ToM (Eq.~\ref{eq:model3}). The point estimate for the $ToM \times Joint$ coefficient is virtually identical across all specifications, suggesting remarkable stability. We find small variation across the width of the credible interval, which varies mechanically with the fixed-effect prior bandwidth. The random-effect shrinkage prior has essentially no effect. This gives further confidence to our estimates of individual ability despite the limited data. Under the strictest shrinkage specification, the estimate is 0.65 [95\% CrI: 0.02, 1.28], with the CI excluding zero. If noisy $\theta$ estimates from sparse measurement were driving the result, heavier shrinkage should have moved the point estimate or dramatically changed the CI, which it did not. Under the widest fixed-effect prior, posterior uncertainty increases and the 95\% interval includes zero while the point estimate is essentially unchanged, indicating that the sensitivity reflects limited user-level data rather than dependence on the prior.
\clearpage

\begin{table}[!th]
\begin{tabular}[t]{llll}
\toprule
Prior Specification & ToM (solo) & ToM $\times$ Joint & SD($\kappa - \theta$)\\
\midrule
Main (paper) & 0.40 [-0.16, 0.99] & 0.65 [0.02, 1.26] & 1.06 [0.78, 1.33]\\
Wider FE & 0.43 [-0.19, 1.06] & 0.64 [-0.06, 1.34] & 1.06 [0.79, 1.33]\\
Tighter FE & 0.37 [-0.11, 0.85] & 0.65 [0.14, 1.15] & 1.06 [0.78, 1.33]\\
Less shrinkage & 0.41 [-0.18, 0.99] & 0.65 [-0.00, 1.29] & 1.09 [0.81, 1.35]\\
More shrinkage & 0.40 [-0.17, 0.97] & 0.65 [0.02, 1.28] & 1.02 [0.74, 1.29]\\
\addlinespace
Wider FE + less shrinkage & 0.42 [-0.21, 1.06] & 0.65 [-0.06, 1.35] & 1.09 [0.81, 1.36]\\
Tighter FE + more shrinkage & 0.36 [-0.11, 0.83] & 0.65 [0.14, 1.15] & 1.02 [0.74, 1.29]\\
Default (\texttt{brms}) & 0.42 [-0.23, 1.08] & 0.65 [-0.07, 1.37] & 1.09 [0.81, 1.37]\\
\bottomrule
\multicolumn{4}{p{.9\linewidth}}{\footnotesize Note. Main paper uses more conservative weakly informative priors with stricter regularization than \texttt{brms} defaults: normal(0, 1) for fixed effects and exponential(1) on SDs.}
\end{tabular}
\vspace{5pt}
\centering
\caption{Sensitivity of ToM interaction estimates to prior specification.
    ToM (solo) is the main effect of ToM in the solo condition.
    ToM $\times$ Joint is the incremental ToM effect in the collaborative condition.
    SD($\kappa - \theta$) is the posterior SD of the collaborative boost random slope.
    All entries show posterior median [95\% CrI].}
\centering
\end{table}

\section{Bivariate Partial Information Decomposition (PID)}\label{sec:PID}

We compute a bivariate partial information decomposition (PID) as a model-free alternative to the IRT analyses reported above. For this we use the answer-first arm of the dataset where users answer the \textit{same question} first alone and then with AI \citep{Chang2025Chatbench}. Because PID requires the joint distribution of sources and target on the same units, the direct-to-AI arm cannot support this analysis as solo and joint items never overlap within an individual. PID allows us to decompose the information provided by two sources about a target into unique information provided by each source, redundant information provided by both, and synergistic information that neither source alone contains \citep{riedl2026emergence}. For each individual–question trial we defined three binary variables: $H$, provisional answer correct; $A$, AI correct on the item (arm-matched few-shot accuracy $> 0.5$ for the AI model the participant was paired with); and $Y$, final answer correct (3,038 trials, 343 users, 396 items). 

We then decompose $I(H,A;Y)$ into redundant, unique, and synergistic atoms using the BROJA measure \citep{bertschinger2014-BROJA}. The implementation is a convex program over distributions preserving the $(H,Y)$ and $(A,Y)$ marginals (minimum-mutual-information is identical to BROJA in our case given the Blackwell-dominated finding below). We report bias-corrected estimates by subtracting the mean of a permutation null ($Y$ shuffled within stratum, $R = 500$); 95\% intervals are from a user-level cluster bootstrap ($B = 1{,}000$). We report atoms for all items and by tercile of solo item difficulty. To initialize the optimizer, we smooth empty cells ($\epsilon = 10^{-6}$). 

The PID and IRT analyses complement each other rather than being redundant. The IRT model estimates person-level latent abilities and question difficulties, propagates uncertainty, and identifies who benefits, on which items, and whether the mechanism is additive; the PID makes no such structural assumptions and instead partitions the information that human and AI carry about the final outcome into unique, redundant and synergistic components, providing a model-free test that the synergistic component is positive.

\begin{figure}[h!]
\centering
\includegraphics[width=1\columnwidth]{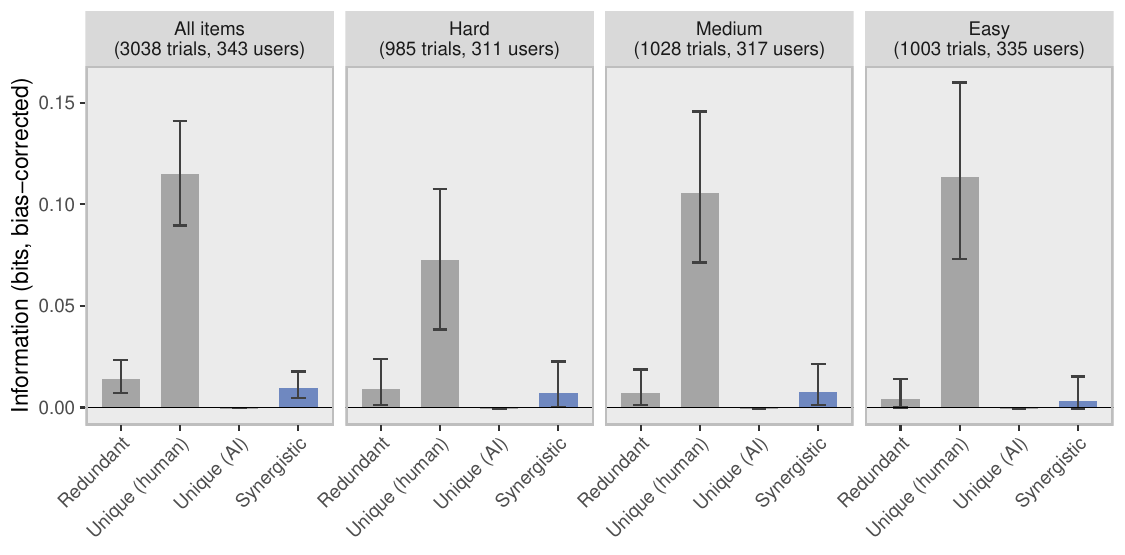}
\caption{
    \textbf{Model-free partial information decomposition of final-answer correctness into Redundant, Unique, and Synergistic information.}
    Bars show information in bits for all questions and by easy/medium/hard terciles (based on solo difficulty). Point estimates are bias-corrected by subtracting the mean of a within-stratum permutation null ($Y$ shuffled, $R = 500$); error bars are 95\% participant-level cluster-bootstrap intervals ($B = 1{,}000$). $\mathrm{Un}(A)$ is zero because the channel $Y \rightarrow A$ is Blackwell-dominated by $Y \rightarrow H$, a property of binary-source PID with a question-level proxy for $A$ rather than evidence that the AI contributes no information.
}
\label{fig:PID}
\end{figure}

We find the synergistic atom is credibly positive overall and in the hard and medium strata (Figure \ref{fig:PID}). Synergy also accounts for a larger share of total information on hard items, consistent with the IRT finding that collaborative gains concentrate where solo performance is weakest. 
This positive synergistic atom is expected under the additive-logit structure recovered in the main IRT analyses. The decomposition is inconsistent with pure substitution, under which the final answer would be a copy of the AI's and $\mathrm{Un}(A)$ would dominate. Instead, the user's initial answer---whether correct or incorrect---strongly dictates the final outcome. $\mathrm{Un}(A)$ is zero because the question-level AI variable is Blackwell-dominated by the provisional answer, so the PID cannot separate the AI's contribution from the participant's. 
Consider this alternative: If the human simply closed their eyes, ignored their initial instinct ($H$), and copied the AI ($A$), then $Y$ would be a copy of $A$. In that scenario, $A$ would Blackwell-dominate $H$, and $\mathrm{Un}(H)$ would be zero. Because $\mathrm{Un}(A) = 0$ and $\mathrm{Un}(H) > 0$, the human is not merely adopting the AI's answer.

Note that using the few-shot accuracy of the AI for $A$ is only a proxy for the AI's contribution on the question. It does not fully capture the AI portion of the dialogue. The dialogue likely carries more information, thus making our findings of synergy a conservative estimate and contributing to the $\mathrm{Un}(A) \sim 0$ estimates. Detecting a credibly positive synergistic atom despite this extreme informational bottleneck provides robust evidence of true collaborative gains.
Because the user decides whether to retain or revise their preliminary answer, information from the AI enters the final outcome only through that decision, which loads it onto $\mathrm{Syn}$ and $\mathrm{Un}(H)$ rather than $\mathrm{Un}(A)$; this pattern is what calibrated reliance predicts \citep{Caplin2025ABC}, in which adoption is conditional on the participant's assessment of their own and the AI's likely correctness (the three-variable decomposition cannot itself separate calibrated adoption from unconditional anchoring).

Because synergy in this decomposition equals $I(A;Y | H)$, synergy would also be positive if users mix keeping and deferring without regard to which source is correct. To rule this out, we examine the trials on which the two disagreed. On trials where the user’s provisional answer and the AI’s answer disagree ($N = 1{,}395$), users kept a correct provisional answer in $p_{10}$ of cases and adopted a correct AI answer in $p_{01}$. Under any policy that keeps or defers without regard to which source is correct, these rates sum to $1$. We observe a sum of $1.47$ ([95\% CI: 1.42, 1.53] using a user-level cluster bootstrap). Allowing for item-level coding of AI correctness, the conservative excess is $0.44$ [95\% CI: 0.36, 0.52] for \gpt~and $0.46$ [95\% CI: 0.36, 0.56] for \llama. This suggests that discerning reliance accounts for at least part of the observed synergy. It does not exclude synergy from additional integration on questions where both sources were wrong.

\section{A Non-Interactive Setting as Negative Control}\label{si:bansal}

\paragraph{Data.} We use the released data of \citet{bansal2021does} available at \url{https://github.com/uw-hai/Complementary-Performance}. Users ($N = 1{,}621$) were randomly assigned to answer all items of one task either alone ($n = 283$) or with a fixed AI that displayed its answer, confidence, and explanation before the participant responded ($n = 1{,}338$). The AI was a fixed classifier fine-tuned for each task, not a chat-based LLM. Its answers and confidence were precomputed and identical for all participants. The tasks were two binary sentiment-classification tasks (book and beer reviews, 50 items each) and one four-option logical-reasoning task (20 LSAT items). The AI was correct on 84\%, 84\% and 65\% of items, respectively, so its correctness is known exactly for every item (wrong on 8, 8 and 7 items). Users in the joint condition were further randomized across several conditions that provided explanations. Because explanations had no effect on accuracy \citep{bansal2021does}, we pool them with equal weight. Because the study used a between-subject design, solo and collaborative ability cannot be estimated for the same person and Eq.~\ref{eq:model1} does not apply.

\paragraph{Where the gain comes from.} Team accuracy (0.912, 0.889, 0.693) exceeds both unaided human accuracy (0.877, 0.815, 0.582) and AI accuracy (0.840, 0.840, 0.650) on all three tasks, as reported in the original paper. Splitting items by AI correctness shows the source of the gain. On items the AI answers correctly, the human-AI team gains 6.3, 11.3, and 24.7 percentage points over humans working alone. On items it answers incorrectly, they lose 10.9, 13.2, and 14.2~pp. Teams fall below human-alone accuracy on all 23 AI-wrong items. We benchmark the realized gain against \emph{oracle selection}: a team that takes the AI's answer whenever it is correct and the answer a human working alone would give otherwise. Specifically, we take the expected accuracy $\frac{1}{J}\sum_j [a_j + (1-a_j),p_j]$, where $a_j$ indicates AI correctness and $p_j$ is the expected accuracy on item $j$ of individuals working alone ($0.969$, $0.978$, $0.855$).
Human-AI teams realize $(\text{team} - \max(\text{human}, \text{AI}))/(\text{oracle} - \max(\text{human}, \text{AI})) = 38\%$, 36\% and 21\% of this available gain. These shares represent an upper bound, since a team that constructs answers neither member would reach alone could exceed the oracle baseline.

\paragraph{Judgment and deference.} To separate individual differences in deference from differences in judgment, we fit an IRT model with crossed random-effects to data in the joint condition (separately for each decision task in the experiment):

\begin{equation}
\operatorname{logit} P(y_{ij} = 1) = \alpha_{c(i)} + \beta s_j + v_j + \zeta_i + s_j \tau_i,
\qquad s_j = \begin{cases} +1 & \text{AI correct on item } j\\ -1 & \text{AI incorrect}\end{cases}
\label{eq:bansal}
\end{equation}

with condition intercepts $\alpha_c$, item effects $v_j \sim \mathcal{N}(0, \sigma_v^2)$, and bivariate normal individual effects $(\zeta_i, \tau_i)$. The random intercept $\zeta_i$ (\emph{judgment}) is an individual's average logit accuracy across AI-correct and AI-incorrect items. The random slope $\tau_i$ (\emph{deference}) is half the difference: individuals with high $\tau_i$ do better when the AI is right and worse when it is wrong. Because every individual answers every item and the AI's correctness is fixed by design, both effects are identified within individuals. As priors we use $\mathcal{N}(0, 1.5)$ on fixed effects, half-$\mathcal{N}(0, 1)$ on standard deviations, and LKJ(1) on the correlation. We again fit the models using \texttt{brms} and all models converge cleanly (four chains of 2,000 post-warmup draws, $\hat R \le 1.009$, no divergent transitions).

The SD of deference is 0.45 [0.36, 0.53], 0.67 [0.58, 0.76] and 0.73 [0.63, 0.83] on the logit scale (book reviews, beer reviews, LSAT; 95\% credible intervals). The SD of judgment is 0.90 [0.81, 1.00], 0.74 [0.65, 0.83] and 0.77 [0.67, 0.88]. Deference therefore varies about half as much as judgment on book reviews and nearly as much on the other two tasks (ratios 0.49, 0.91, 0.94). Raw accuracy comparisons cannot separate the two, because individuals' overall accuracy mixes both and depends on the share of items the AI gets right.

\paragraph{Does complementarity generalize to the task?} When Bansal et al.~show that human-AI teams beat their best member they treat the item set as fixed. However, items in their tasks---and LSAT in particular---are designed as interchangeable measures of general reasoning ability. Items should be treated as a sample from a population of items. Would the result hold if Bansal et al.~had picked different questions? That is, do the complementarity claims in \citet{bansal2021does} generalize to the tasks the items are meant to represent? This is something that our crossed random-effects IRT model does by construction. 
Using this approach, we find that the finding that human-AI teams beat their best member generalizes on both sentiment tasks but \textit{not} on LSAT.

Specifically, we fit one crossed random-effects model per task to both arms. It has cell means for each condition $\times$ AI correctness, correlated item effects for the solo and joint conditions, and individual intercepts and slopes as above (we allow separate covariances per arm). From the model posteriors we compute population accuracies, integrating over individuals. Item uncertainty is propagated by a Bayesian bootstrap over the estimated item effects: Dirichlet weights within each AI-correctness stratum, so the AI's accuracy stays at its design value. We use this rather than a normal item population because item difficulties are skewed and the normal approximation misstates mean accuracy. The human-AI team's advantage over the better of human and AI remains on both sentiment tasks (+3.6~pp [1.0, 6.6]; +4.8~pp [1.0, 7.2]) but not on LSAT (+4.2~pp [$-2.1$, 9.9]). LSAT has only 20 items, 7 of them AI-incorrect, so its advantage over the AI does not generalize beyond these items. The losses on AI-incorrect items hold on all three tasks ($-10.6$, $-12.9$, $-14.1$~pp; all intervals exclude zero).

\section{Three-Arm Structural Mechanism Model ($2 \times 2$ User–Item Interactions)} \label{si:mechanism}

\paragraph{Data Structure and Multi-Arm Design.}
To test how human skills interact with item-level human difficulty and AI capability we develop a joint model that combines data from three experimental arms. In addition to the solo and joint data from above, we also use the standalone AI benchmark data. To summarize all three arms in one place:

\begin{itemize}
    \item \textbf{Human Solo Arm (solo):} Binary trial-level accuracy $y_{ij} \in \{0,1\}$ for individual $i$ answering item $j$ unaided in Phase 1.
    \item \textbf{Standalone AI Benchmark Arm (aiq):} Binomial accuracy $y_{jm} \in \{0, \dots, 50\}$ across $n_{jm} = 50$ independent stochastic evaluations of item $j$ by AI model $m \in \{\text{GPT-4o}, \text{Llama}\}$ (free text accuracy). Evaluating each item 50 times per model in free-text format directly measures continuous item-specific AI capability on an unbounded log-odds scale---avoiding multiple-choice guessing floors---rather than thresholding binary solvability as in \supp~\ref{sec:substitution} and \ref{sec:PID}.
    \item \textbf{Human–AI Collaboration Arm (joint, direct-to-AI condition only):} Binary trial-level accuracy $y_{ijm} \in \{0,1\}$ for individual $i$ collaborating with AI model $m$ on a set of items $j$ different from those the user answered in the solo condition.
\end{itemize}

\paragraph{Hierarchical Bilinear Specification.} Let $p$ denote the probability of a correct response, with logistic link $\text{logit}(p) = \log(p / (1 - p))$. Across the three arms, the linear predictor is specified jointly as:

\begin{equation}    
\begin{aligned} \text{Solo:} \quad \text{logit}(p_{ij}^{\text{solo}}) &= c_{\text{solo}} + \theta_i - d_j, \\[4pt] \text{AI-only:} \quad \text{logit}(p_{jm}^{\text{ai}}) &= c_{\text{ai},m} + \eta_{jm}, \\[4pt] \text{Joint:} \quad \text{logit}(p_{ijm}^{\text{joint}}) &= c_{\text{joint},m} + b_1 \theta_i + b_2 \eta_{jm} - b_4 d_j + b_0 \kappa^\perp_i + v_j \\ &\quad + b_3 (\theta_i \eta_{jm}) + b_5 (\kappa^\perp_i \eta_{jm}) + b_7 (\theta_i d_j) + b_8 (\kappa^\perp_i d_j). \end{aligned}
\label{eq:2x2}
\end{equation}

Individual random effects follow independent normal distributions:

$$
\theta_i \sim \mathcal{N}(0, \sigma_\theta^2), \qquad \kappa^\perp_i \sim \mathcal{N}(0, 1), \qquad \theta_i \perp \kappa^\perp_i,
$$

with $b_0 \ge 0$. Item random effects allow arbitrary correlation across solo human difficulty $d_j$ and model-specific AI capability $(\eta_{j,\text{GPT}}, \eta_{j,\text{Llama}})$, alongside an independent residual item perturbation $v_j$ capturing collaboration-specific item idiosyncrasies:

$$
\begin{pmatrix} d_j \\ \eta_{j,\text{GPT}} \\ \eta_{j,\text{Llama}} \end{pmatrix} \sim \mathcal{N}\!\left(\mathbf{0}, \Sigma_{\text{item}}\right), \qquad v_j \sim \mathcal{N}(0, \sigma_v^2).
$$

\paragraph{Structural Identification and Relation to Model \ref{eq:model1}.}
In a standalone collaboration dataset, interacting two latent user traits with two latent item traits leaves factor orientations underidentified up to rotation. Our three-arm within-subject design resolves this indeterminacy through external measurement anchoring.

\textbf{Item axes ($d_j, \eta_{jm}$):} The solo arm pins down human item difficulty $d_j$ (in unit Rasch scale relative to $\theta_i$), while the 50-trial AI-only arm pins down each AI model's item-level log-odds signal $\eta_{jm}$ and the full cross-arm item covariance matrix $\Sigma_{\text{item}}$. Empirically, items that are harder for humans are also harder for AI ($\text{Corr}(d_j, \eta_{j,\text{GPT}}) = -0.40$; $\text{Corr}(d_j, \eta_{j,\text{Llama}}) = -0.58$). Including both $\eta_{jm}$ and $d_j$ in the $2 \times 2$ interaction block ensures that individual sensitivity to AI capability ($b_3, b_5$) is not confounded by omitted interactions with intrinsic human difficulty ($b_7, b_8$).

\textbf{User axes ($\theta_i, \kappa^\perp_i$):} Because each subject is observed in both the solo and joint phases, the solo data externally anchors solo domain skill $\theta_i$. Whereas Eq.~\ref{eq:model1} in the main text estimates unresidualized solo and collaborative abilities and shows they correlate only moderately, here we apply a triangular (Cholesky) decomposition that projects total collaborative skill onto $\theta_i$ and defines $\kappa^\perp_i$ as \textit{solo-orthogonal collaboration skill} ($\text{Cov}(\theta_i, \kappa^\perp_i) = 0$). Fixing $\text{SD}(\kappa^\perp_i) = 1$ with sign normalization $b_0 > 0$ avoids collinearity in the interaction terms while ensuring all parameters remain finite even if baseline or interaction loadings approach zero. Under this decomposition, $b_1, b_3, b_7$ capture the total linear projection on solo ability $\theta_i$, whereas $b_0, b_5, b_8$ isolate variance strictly orthogonal to solo ability.

\paragraph{Bayesian Estimation, Priors, and Robustness Tests.}
As before we estimate models with $\mathrm{brms}$ (8 chains, 8,000 iterations per chain, 2,000 warmup). We placed weakly informative priors on all parameters: $\mathcal{N}(0, 2)$ on human/joint intercepts ($c_{\text{solo}}, c_{\text{joint},m}$), $\mathcal{N}(0, 5)$ on standalone AI intercepts ($c_{\text{ai},m}$, accommodating high baseline \gpt~accuracy), $\mathcal{N}(1, 1)$ on main-effect transmission slopes ($b_1, b_2, b_4$), truncated normal $\mathcal{N}_+(0.5, 1)$ on $b_0$, symmetric $\mathcal{N}(0, 1)$ priors on all four bilinear interaction coefficients ($b_3, b_5, b_7, b_8$), half-normal $\mathcal{N}_+(0, 1)$ priors on $\sigma_\theta, \sigma_d, \sigma_v$, half-Student-$t_3(0, 5)$ priors on $\sigma_{\eta,\text{GPT}}$ and $\sigma_{\eta,\text{Llama}}$, and an $\text{LKJ}(2)$ prior on the item correlation matrix. All chains converged cleanly with zero divergent transitions.

To verify that interaction estimates in the $\mathrm{joint}$ arm do not reflect unmodeled two-parameter logistic (2PL) item discrimination in $\mathrm{solo}$ performance, we estimated an extended specification allowing $\theta_i \times d_j$ and $\theta_i \times \eta_j$ interactions in the solo equation; both solo discrimination terms were near zero and their CrIs include zero and all substantive conclusions remained virtually identical.

\textbf{OR-Baseline.} To distinguish selective answer aggregation from co-construction, we use posterior draws from the three-arm model to compute, for every observed joint trial, the same user's predicted solo accuracy ($p^{\text{solo}}_{ij}$), the paired model's standalone accuracy ($p^{\text{ai}}_{jm}$), and the predicted collaborative accuracy ($p^{\text{joint}}_{ijm}$). We compare mean collaborative accuracy against three benchmarks, each averaged over joint trials: the best single source ($B_0 = \max\big(\operatorname{mean}\, p^{\text{solo}}_{ij},\ \operatorname{mean}\, p^{\text{ai}}_{jm}\big)$; exceeding it constitutes complementary team performance in the sense of \citet{bansal2021does}); item-level calibrated deference ($B_1 = \operatorname{mean}\, \max(p^{\text{solo}}_{ij}, p^{\text{ai}}_{jm})$); and the oracle truth-wins OR ceiling ($B_2 = \operatorname{mean}\, [1 - (1 - p^{\text{solo}}_{ij})(1 - p^{\text{ai}}_{jm})]$), which upper-bounds any pure answer-selection rule under conditional independence. Across posterior draws, we evaluate $p^{\text{joint}} - B_1$, $p^{\text{joint}} - B_2$, and the normalized synergy index $(p^{\text{joint}} - B_0)/(B_2 - B_0)$ (0 at the best single source, 1 at the OR ceiling) overall, by AI model, within ``hard-for-both'' cells ($\bar{p}^{\text{solo}}_{ij} < 0.5$ and $\bar{p}^{\text{ai}}_{jm} < 0.5$), and at a +1\,SD $\kappa^{\perp}_i$ counterfactual collaboration skill level. Values within $B_1 < p^{\text{joint}} \le B_2$ indicate calibrated answer-level signal extraction, whereas $p^{\text{joint}} > B_2$ is required as evidence of co-constructing solutions that neither partner produces alone.

\end{document}